\documentstyle[11pt,fleqn,psfig]{article}
 
\def\diatop[#1|#2]{{\setbox1=\hbox{{#1{}}}\setbox2=\hbox{{#2{}}}%
                    \dimen0=\ifdim\wd1>\wd2\wd1\else\wd2\fi%
                    \dimen1=\ht2\advance\dimen1by-1ex%
                    \setbox1=\hbox to1\dimen0{\hss#1\hss}%
                    \rlap{\raise1\dimen1\box1}%
                    \hbox to1\dimen0{\hss#2\hss}}}%
 
\def\ipa{\ipatwelverm}
 
\def\inva{\edef\next{\the\font}\ipa\char'000\next}%
\def\scripta{\edef\next{\the\font}\ipa\char'001\next}%
\def\nialpha{\edef\next{\the\font}\ipa\char'002\next}%
\def\invscripta{\edef\next{\the\font}\ipa\char'003\next}%
\def\invv{\edef\next{\the\font}\ipa\char'004\next}%
 
\def\crossb{\edef\next{\the\font}\ipa\char'005\next}%
\def\barb{\edef\next{\the\font}\ipa\char'006\next}%
\def\slashb{\edef\next{\the\font}\ipa\char'007\next}%
\def\hookb{\edef\next{\the\font}\ipa\char'010\next}%
\def\nibeta{\edef\next{\the\font}\ipa\char'011\next}%
 
\def\slashc{\edef\next{\the\font}\ipa\char'012\next}%
\def\curlyc{\edef\next{\the\font}\ipa\char'013\next}%
\def\clickc{\edef\next{\the\font}\ipa\char'014\next}%
 
\def\crossd{\edef\next{\the\font}\ipa\char'015\next}%
\def\bard{\edef\next{\the\font}\ipa\char'016\next}%
\def\slashd{\edef\next{\the\font}\ipa\char'017\next}%
\def\hookd{\edef\next{\the\font}\ipa\char'020\next}%
\def\taild{\edef\next{\the\font}\ipa\char'021\next}%
\def\dz{\edef\next{\the\font}\ipa\char'022\next}%
\def\eth{\edef\next{\the\font}\ipa\char'023\next}%
\def\scd{\edef\next{\the\font}\ipa\char'024\next}%
 
\def\schwa{\edef\next{\the\font}\ipa\char'025\next}%
\def\er{\edef\next{\the\font}\ipa\char'026\next}%
\def\reve{\edef\next{\the\font}\ipa\char'027\next}%
\def\niepsilon{\edef\next{\the\font}\ipa\char'030\next}%
\def\revepsilon{\edef\next{\the\font}\ipa\char'031\next}%
\def\hookrevepsilon{\edef\next{\the\font}\ipa\char'032\next}%
\def\closedrevepsilon{\edef\next{\the\font}\ipa\char'033\next}%
 
\def\scriptg{\edef\next{\the\font}\ipa\char'034\next}%
\def\hookg{\edef\next{\the\font}\ipa\char'035\next}%
\def\scg{\edef\next{\the\font}\ipa\char'036\next}%
\def\nigamma{\edef\next{\the\font}\ipa\char'037\next}
\def\ipagamma{\edef\next{\the\font}\ipa\char'040\next}%
\def\babygamma{\edef\next{\the\font}\ipa\char'041\next}%
 
\def\hv{\edef\next{\the\font}\ipa\char'042\next}%
\def\crossh{\edef\next{\the\font}\ipa\char'043\next}%
\def\hookh{\edef\next{\the\font}\ipa\char'044\next}%
\def\hookheng{\edef\next{\the\font}\ipa\char'045\next}%
\def\invh{\edef\next{\the\font}\ipa\char'046\next}%
 
\def\bari{\edef\next{\the\font}\ipa\char'047\next}%
\def\dlbari{\edef\next{\the\font}\ipa\char'050\next}
\def\niiota{\edef\next{\the\font}\ipa\char'051\next}%
\def\sci{\edef\next{\the\font}\ipa\char'052\next}%
\def\barsci{\edef\next{\the\font}\ipa\char'053\next}
 
\def\invf{\edef\next{\the\font}\ipa\char'054\next}%
 
\def\tildel{\edef\next{\the\font}\ipa\char'055\next}%
\def\barl{\edef\next{\the\font}\ipa\char'056\next}%
\def\latfric{\edef\next{\the\font}\ipa\char'057\next}%
\def\taill{\edef\next{\the\font}\ipa\char'060\next}%
\def\lz{\edef\next{\the\font}\ipa\char'061\next}%
\def\nilambda{\edef\next{\the\font}\ipa\char'062\next}%
\def\crossnilambda{\edef\next{\the\font}\ipa\char'063\next}%
 
\def\labdentalnas{\edef\next{\the\font}\ipa\char'064\next}%
\def\invm{\edef\next{\the\font}\ipa\char'065\next}%
\def\legm{\edef\next{\the\font}\ipa\char'066\next}%
 
\def\nj{\edef\next{\the\font}\ipa\char'067\next}%
\def\eng{\edef\next{\the\font}\ipa\char'070\next}%
\def\tailn{\edef\next{\the\font}\ipa\char'071\next}%
\def\scn{\edef\next{\the\font}\ipa\char'072\next}%
 
\def\clickb{\edef\next{\the\font}\ipa\char'073\next}%
\def\baro{\edef\next{\the\font}\ipa\char'074\next}%
\def\openo{\edef\next{\the\font}\ipa\char'075\next}%
\def\niomega{\edef\next{\the\font}\ipa\char'076\next}%
\def\closedniomega{\edef\next{\the\font}\ipa\char'077\next}%
\def\oo{\edef\next{\the\font}\ipa\char'100\next}%
 
\def\barp{\edef\next{\the\font}\ipa\char'101\next}%
\def\thorn{\edef\next{\the\font}\ipa\char'102\next}%
\def\niphi{\edef\next{\the\font}\ipa\char'103\next}%
 
\def\flapr{\edef\next{\the\font}\ipa\char'104\next}%
\def\legr{\edef\next{\the\font}\ipa\char'105\next}%
\def\tailr{\edef\next{\the\font}\ipa\char'106\next}%
\def\invr{\edef\next{\the\font}\ipa\char'107\next}%
\def\tailinvr{\edef\next{\the\font}\ipa\char'110\next}%
\def\invlegr{\edef\next{\the\font}\ipa\char'111\next}%
\def\scr{\edef\next{\the\font}\ipa\char'112\next}%
\def\invscr{\edef\next{\the\font}\ipa\char'113\next}%
 
\def\tails{\edef\next{\the\font}\ipa\char'114\next}%
\def\esh{\edef\next{\the\font}\ipa\char'115\next}%
\def\curlyesh{\edef\next{\the\font}\ipa\char'116\next}%
\def\nisigma{\edef\next{\the\font}\ipa\char'117\next}%
 
\def\tailt{\edef\next{\the\font}\ipa\char'120\next}%
\def\tesh{\edef\next{\the\font}\ipa\char'121\next}%
\def\clickt{\edef\next{\the\font}\ipa\char'122\next}%
\def\nitheta{\edef\next{\the\font}\ipa\char'123\next}%
 
\def\baru{\edef\next{\the\font}\ipa\char'124\next}%
\def\slashu{\edef\next{\the\font}\ipa\char'125\next}%
\def\niupsilon{\edef\next{\the\font}\ipa\char'126\next}%
\def\scu{\edef\next{\the\font}\ipa\char'127\next}%
\def\barscu{\edef\next{\the\font}\ipa\char'130\next}%
 
\def\scriptv{\edef\next{\the\font}\ipa\char'131\next}%
 
\def\invw{\edef\next{\the\font}\ipa\char'132\next}%
 
\def\nichi{\edef\next{\the\font}\ipa\char'133\next}%
 
\def\invy{\edef\next{\the\font}\ipa\char'134\next}%
\def\scy{\edef\next{\the\font}\ipa\char'135\next}%
 
\def\curlyz{\edef\next{\the\font}\ipa\char'136\next}%
\def\tailz{\edef\next{\the\font}\ipa\char'137\next}%
\def\yogh{\edef\next{\the\font}\ipa\char'140\next}%
\def\curlyyogh{\edef\next{\the\font}\ipa\char'141\next}%
 
\def\glotstop{\edef\next{\the\font}\ipa\char'142\next}%
\def\revglotstop{\edef\next{\the\font}\ipa\char'143\next}%
\def\invglotstop{\edef\next{\the\font}\ipa\char'144\next}%
\def\ejective{\edef\next{\the\font}\ipa\char'145\next}%
\def\reveject{\edef\next{\the\font}\ipa\char'146\next}%
 
\def\upt{\edef\next{\the\font}\ipa\char'154\next}
\def\downt{\edef\next{\the\font}\ipa\char'155\next}%
\def\leftt{\edef\next{\the\font}\ipa\char'156\next}%
\def\rightt{\edef\next{\the\font}\ipa\char'157\next}%
 
\def\upp{\edef\next{\the\font}\ipa\char'164\next}
\def\downp{\edef\next{\the\font}\ipa\char'165\next}%
\def\leftp{\edef\next{\the\font}\ipa\char'166\next}%
\def\rightp{\edef\next{\the\font}\ipa\char'167\next}%
 
\def\stress{\edef\next{\the\font}\ipa\char'150\next}
\def\secstress{\edef\next{\the\font}\ipa\char'151\next}
 
\def\syllabic{\edef\next{\the\font}\ipa\char'152\next}
 
\def\corner{\edef\next{\the\font}\ipa\char'153\next}%
 
\def\halflength{\edef\next{\the\font}\ipa\char'160\next}
\def\length{\edef\next{\the\font}\ipa\char'161\next}
 
\def\underdots{\edef\next{\the\font}\ipa\char'162\next}%
 
\def\ain{\edef\next{\the\font}\ipa\char'163\next}
 
\def\overring{\edef\next{\the\font}\ipa\char'170\next}%
\def\underring{\edef\next{\the\font}\ipa\char'171\next}%
 
\def\open{\edef\next{\the\font}\ipa\char'172\next}%
 
\def\midtilde{\edef\next{\the\font}\ipa\char'173\next}%
\def\undertilde{\edef\next{\the\font}\ipa\char'174\next}%
 
\def\underwedge{\edef\next{\the\font}\ipa\char'175\next}%
 
\def\polishhook{\edef\next{\the\font}\ipa\char'176\next}%

\font\ipatenrm=wsuipa10

\onecolumn
\begin{document}

\bibliographystyle{plain}

\begin{titlepage}
\advance\topmargin by 0.5in
\begin{center}
\vspace{1.0in}
{\large MASSACHUSETTS INSTITUTE OF TECHNOLOGY\\}
\vspace{.05in}
{\large ARTIFICIAL INTELLIGENCE LABORATORY\\}
\vspace{.05in}
{\large and\\}
\vspace{.05in}
{\large CENTER FOR BIOLOGICAL AND COMPUTATIONAL LEARNING\\}
\vspace{.05in}
{\large DEPARTMENT OF BRAIN AND COGNITIVE SCIENCES}
\end{center}
\vspace{.2in}
{\large A.I.\ Memo No.\ 1558 \hfill November, \number\year\\}
{\large C.B.C.L.\ Memo No.\ 129}
\vspace{.2in}
\begin{center}

  {\Large\bf The Unsupervised Acquisition of a Lexicon from\\ Continuous
    Speech} \\ 

\bigskip
\bigskip
{\large\bf Carl de Marcken\\{\tt cgdemarc@ai.mit.edu}}\\ \bigskip

{\normalsize This publication can be retrieved by anonymous ftp to
publications.ai.mit.edu.}

\end{center}
\vfill
\begin{abstract}
  {\small\noindent
We present an unsupervised learning algorithm that
acquires a natural-language lexicon from raw speech.  The algorithm is
based on the optimal encoding of symbol sequences in an MDL framework,
and uses a hierarchical representation of language that overcomes many
of the problems that have stymied previous grammar-induction
procedures.  The forward mapping from symbol sequences to the speech
stream is modeled using features based on articulatory gestures.  We
present results on the acquisition of lexicons and language models from
raw speech, text, and phonetic transcripts, and demonstrate that our
algorithm compares very favorably to other reported results with
respect to segmentation performance and statistical efficiency.
\par}

\end{abstract}
\vfill
\begin{center} \footnotesize
Copyright \copyright\ Massachusetts Institute of Technology, 1995
\end{center}
\vfill {\footnotesize \noindent This report describes research done at the
  Artificial Intelligence Laboratory of the Massachusetts Institute of
  Technology.  This research is supported by NSF grant 9217041-ASC and ARPA
  under the HPCC and AASERT programs. \par}
\vspace{.5in}
\end{titlepage}

\newcommand{\pageline}{\vspace{-.1in}\underline{\hspace{\columnwidth}}}
\newcommand{\kwl}{{k\stackrel{w}{\rightarrow}l}}
\newcommand{\lwk}{{k\stackrel{w}{\rightarrow}l}}
\newcommand{\wordsumkwl}{{\textstyle\sum_{\kwl}}}
\newcommand{\wordsumlwk}{{\textstyle\sum_{\lwk}}}
\newcommand{\subseq}[3]{{_{#1}{#2}_{#3}}}
\newcommand{\unib}{b}
\newcommand{\unid}{d}
\newcommand{\unig}{g}
\newcommand{\unip}{p}
\newcommand{\unit}{t}
\newcommand{\unik}{k}
\newcommand{\uniJ}{\v{j}}
\newcommand{\uniC}{\v{c}}
\newcommand{\unis}{s}
\newcommand{\uniS}{\v{s}}
\newcommand{\uniz}{z}
\newcommand{\uniZ}{\v{z}}
\newcommand{\unif}{f}
\newcommand{\uniT}{\nitheta}
\newcommand{\univ}{v}
\newcommand{\uniD}{\eth}
\newcommand{\unim}{m}
\newcommand{\unin}{n}
\newcommand{\uniG}{\eng}
\newcommand{\unil}{l}
\newcommand{\unir}{r}
\newcommand{\uniw}{w}
\newcommand{\uniy}{y}
\newcommand{\unih}{h}
\newcommand{\uniH}{\hookh}
\newcommand{\unii}{i}
\newcommand{\uniI}{\sci}
\newcommand{\uniE}{\niepsilon}
\newcommand{\unie}{e}
\newcommand{\uniA}{\ae}
\newcommand{\unia}{a}
\newcommand{\uniAH}{\invv} 
\newcommand{\uniO}{\openo}
\newcommand{\unio}{o}
\newcommand{\uniU}{\niupsilon}
\newcommand{\uniu}{u}
\newcommand{\uniAX}{\schwa} 
\newcommand{\uniIX}{\bari} 
\newcommand{\uniEPI}{-} 

\section{Introduction}

Internally, a sentence is a sequence of discrete elements drawn from a
finite vocabulary.  Spoken, it becomes a continuous signal-- a series of
rapid pressure changes in the local atmosphere with few obvious divisions.
How can a pre-linguistic child, or a computer, acquire the skills necessary
to reconstruct the original sentence?  Specifically, how can it learn the
vocabulary of its language given access only to highly variable, continuous
speech signals? We answer this question, describing an algorithm that
produces a linguistically meaningful lexicon from a raw speech stream.  Of
course, it is not the first answer to how an utterance can be segmented and
classified given a fixed vocabulary, but in this work we are specifically
concerned with the {\em unsupervised acquisition of a lexicon, given no
  prior language-specific knowledge}.

In contrast to several prior proposals, our algorithm makes no assumptions
about the presence of facilitative side information, or of cleanly spoken
and segmented speech, or about the distribution of sounds within words.  It
is instead based on optimal coding in a minimum description length (MDL)
framework.  Speech is encoded as a sequence of articulatory feature
bundles, and compressed using a hierarchical dictionary-based coding
scheme.  The optimal dictionary is the one that produces the shortest
description of both the speech stream and the dictionary itself.  Thus, the
principal motivation for discovering words and other facts about language
is that this knowledge can be used to improve compression, or equivalently,
prediction.

The success of our method is due both to the representation of language we
adopt, and to our search strategy.  In our hierarchical encoding scheme,
all linguistic knowledge is represented in terms of other linguistic
knowledge.  This provides an incentive to learn as much about the general
structure of language as possible, and results in a prior that serves to
discriminate against words and phrases with unnatural structure.  The
search and parsing strategies, on the other hand, deliberately avoid
examining the internal representation of knowledge, and are therefore not
tied to the history of the search process.  Consequently, the algorithm is
relatively free to restructure its own knowledge, and does not suffer from
the local-minima problems that have plagued other grammar-induction
schemes.


At the end, our algorithm produces a lexicon, a statistical language model,
and a segmentation of the input.  Thus, it has diverse application in
speech recognition, lexicography, text and speech compression, machine
translation, and the segmentation of languages with continuous orthography.
This paper presents acquisition results from text and phonetic transcripts,
and preliminary results from raw speech.  So far as we know, these are the
first reported results on learning words directly from speech without prior
knowledge.  Each of our tests is on complex input: the TIMIT speech
collection, the Brown text corpus~\cite{Francis82}, and the CHILDES
database of mothers' speech to children \cite{MacWhinney85}.  The final
words and segmentations accord well with our linguistic intuitions (this is
quantified), and the language models compare very favorably to other
results with respect to statistical efficiency.  Perhaps more
importantly, the work here demonstrates that supervised training is {\em
  not necessary} for the acquisition of much of language, and offers
researchers investigating the acquisition of syntax and other higher
processes a firm foundation to build up from.

The remainder of this paper is divided into nine sections, starting with the
problem as we see it (\ref{problem}) and the learning framework we attack
the problem from (\ref{framework}).  (\ref{speechmodel}) explains how we
link speech to the symbolic representation of language described in
(\ref{grammar}). (\ref{search}) is about our search algorithm,
(\ref{results}) contains results, and (\ref{extensions}) discusses how the
learning framework extends to the acquisition of word meanings and syntax.
Finally, (\ref{related}) frames the work in relation to previous work and
(\ref{conclusions}) concludes.

\section{The Problem}\label{problem}

Broadly, the task we are interested in is this: a listener is presented
with a lengthy but finite sequence of utterances.  Each utterance is an
acoustic signal, sensed by an ear or microphone, and may be paired with
information perceived by other senses.  From these signals, the listener
must acquire the necessary expertise to map a novel utterance into a
representation suitable for higher analysis, which we will take to be a
sequence of words drawn from a known lexicon.

To make this a meaningful problem, we should adopt some objective
definition of what it means to be a word that can be used to evaluate
results.  Unfortunately, there is no single useful definition of a word:
the term encompasses a diverse collection of phenomena that seems to vary
substantially from language to language (see Spencer~\cite{Spencer91}).
For instance, {\it wanna} is a single phonological word, but at the level
of syntax is best analyzed as {\it want to}.  And while {\it common cold}
may on phonological, morphological and syntactic grounds be two words, for
the purposes of machine translation or the acquisition of lexical
semantics, it is more conveniently treated as a unit.

A solution to this conundrum is to avoid distinguishing between different
levels of representation altogether: in other words, to try to capture as
many linguistically important generalizations as possible without labeling
what particular branch of linguistics they fall into.  This approach
accords well with traditional theories of morphology that assume words have
structure at many levels (and in particular with theories that suppose word
formation to obey similar principles to syntax, see Halle and
Marantz~\cite{Halle93b}).  Peeking ahead, after analyzing
the Brown corpus our algorithm parses the phrase {\it the government of the
  united states} as a single entity, a ``word'' that has a representation.
The components of that representation are also words with representations.
The top levels of this tree structure\footnote{The important aspect of this
  representation is that linguistic knowledge (words, in this case) is
  represented in terms of other knowledge.  It is only for computational
  convenience that we choose such a simple concatenative model of
  decomposition.  At the risk of more complex estimation and parsing
  procedures, it would be possible to choose primitives that combine in
  more interesting ways; see, for example, the work of Della Pietra, Della
  Pietra and Lafferty~\cite{DellaPietra95}.  Certainly there is plenty of
  evidence that a single tree structure is incapable of explaining all the
  interactions between morphology and phonology (so-called {\em bracketing
    paradoxes} and the non-concatenative morphology of the semitic
  languages are well-known examples; see 
  Kenstowicz~\cite{Kenstowicz94} for more).} look like

\begin{center}
  \verb*|[[[ the][ [[govern][ment]]]][[ of][[ the][[ united][[ state]s]]]]]|.
\end{center}

\noindent Here, each unit enclosed in brackets (henceforth these will
simply be called words) has an entry in the lexicon.  There is a word {\em
  united states} that might be assigned a meaning independently of {\em
  united} or {\em states}.  Similarly, if the pronunciation of {\em
  government} is not quite the concatenation of {\em govern} and {\em ment}
(as {\em wanna} is not the concatenation of {\em want} and {\em to})
then there is a level of representation where this is naturally captured.
We submit that this hierarchical representation is considerably more useful
than one that treats {\em the government of the united states} as an atom, or that
provides no structure beyond that obvious from the placement of spaces.

If we accept this sort of representation as an intelligent goal, then why
is it hard to achieve?  First of all, notice that even given a known
vocabulary, continuous speech recognition is very difficult.  Pauses are
rare, as anybody who has listened to a conversation in an unknown language
can attest.  What is more, during speech production sounds blend across
word boundaries, and words undergo tremendous phonological and acoustic
variation: {\em what are you doing} is often pronounced
/\uniw\uniAH\uniC\uniAX\unid\uniu\uniIX\unin/.\footnote{Appendix~A provides
  a table of the phonetic symbols used in this paper, and their
  pronunciations.} Thus, before reaching the language learner, unknown
sounds from an unknown number of words drawn from an unknown distribution
are smeared across each other and otherwise corrupted by various noisy
channels.  From this, the learner must deduce the parameters of the
generating process.  This may seem like an impossible task- after all,
every utterance the listener hears could be a new word.  But if the
listener is a human being, he or she is endowed with a tremendous knowledge
about language, be it in the form of a learning algorithm or a universal
grammar, and this constrains and directs the learning process; some part of
this knowledge pushes the learner to establish equivalence classes over
sounds.  The performance of any machine learning algorithm on this problem
is largely dependent on how well it mimics that behavior.

\section{The Learning Framework}\label{framework}

Traditionally, language acquisition has been viewed as the problem of
finding any grammar\footnote{Throughout this paper, the word {\em grammar}
  refers to a grammar in the formal sense, and not to syntax.} consistent with
a sequence of utterances, each labeled grammatical or ungrammatical.
Gold~\cite{Gold67} discusses the difficulty of this problem at length, in
the context of converging on such a grammar in the limit of arbitrarily
long example sequences.  More than 40 years ago, Chomsky saw the problem
similarly, but aware that there might be many consistent grammars, wrote

\begin{quote}
  In applying this theory to actual linguistic material, we must construct
  a grammar of the proper form\ldots\ Among all grammars meeting this
  condition, we select the simplest.  The measure of simplicity must be
  defined in such a way that we will be able to evaluate directly the
  simplicity of any proposed grammar\ldots\ It is tempting, then, to
  consider the possibility of devising a notational system which converts
  considerations of simplicity into considerations of length.
 ~\cite{Chomsky55}
\end{quote}

A practical framework for natural language learning that retains Chomsky's
intuition about the quality of a grammar being inversely related to its
length and avoids the pitfalls of the grammatical vs.\ ungrammatical
distinction is that of {\em stochastic complexity}, as embodied in the
minimum description length (MDL) principle of
Rissanen~\cite{Rissanen78,Rissanen89}.  This principle can be interpreted
as follows: a theory is a model of some process, correct only in so much as
it reliably predicts the outcome of that process.  In comparing theories,
performance must be weighed against complexity-- a baroque theory can
explain any data, but is unlikely to generalize well.  The best theory is
the simplest one that adequately predicts the observed outcome of the
process.  This notion of optimality is formalized in terms of the combined
length of an encoding of both the theory and the data: the best theory is
the one with the shortest such encoding. Put again in terms of language, a
grammar $G$ is a stochastic theory of language production that assigns a
probability $p_{G}(u)$ to every conceivable utterance $u$.  These
probabilities can be used to design an efficient code for utterances;
information theory tells us that $u$ can be encoded using $-\log p_{G}(u)$
bits.\footnote{We assume a minimal familiarity with information theory; see
  Cover and Thomas ~\cite{Cover91} for an introduction.} Therefore, if $U$
is a set of utterances and $|G|$ is the length of the shortest description
of $G$, the combined description length of $U$ and $G$ is $|G| + \sum_{u\in
  U}-\log p_{G}(u)$.  The best grammar for $U$ is the one that minimizes
this quantity.  The process of minimizing it is equivalent to optimally
compressing $U$.

We adopt this MDL framework.  It is well-defined, has a foundation in
information complexity, and (as we will see) leads directly to a convenient
lexical representation.  For our purposes, we choose the class of grammars
in such a way that each grammar is essentially a lexicon.  It is our
premise that within this class, the grammar with the best predictive
properties (the shortest description length) is the lexicon of the source.
Additionally, the competition to compress the input provides a noble
incentive to learn more about the source language than just the lexicon,
and to make use of all cross-linguistic invariants.

We are proposing to learn a lexicon indirectly, by minimizing a function of
the input (the description length) that is parameterized over the lexicon.
This is a risky strategy; certainly a supervised framework would be more
likely to succeed.  Historically, there has been a large community
advocating the view that child language acquisition relies on supervisory
information, either in the form of negative feedback after ungrammatical
productions, or clues present in the input signal that transparently encode
linguistic structure.  The supporting argument has always been that of last
resort: supervised training may explain how learning is possible.  Gold's
proof~\cite{Gold67} that most powerful classes of formal languages are
unlearnable without both positive and negative examples is cited as
additional evidence for the necessity of side information.  Sokolov and
Snow~\cite{Sokolov94} discuss this further and survey arguments that implicit
negative evidence is present in the learning environment.  Along the same
lines, several researchers, notably Jusczyk~\cite{Jusczyk93,Jusczyk94} and
Cutler~\cite{Cutler94}, argue that there are clues in the speech signal,
such as prosody, stress and intonation patterns, that can be used to
segment the signal into words, and that children do in fact attend to these
clues.  Unfortunately, the clues are almost always language specific, which
merely shifts the question to how the clues are acquired.

We prefer to leave open the question of whether children make use of
supervisory information, and attack the question of whether such
information is {\em necessary} for language acquisition.  Gold's proof, for
example, does not hold for suitably constrained classes of languages or for
grammars interpreted in a probabilistic framework.  Furthermore, both the
acquisition and use of prosodic and intonational clues for segmentation
falls out naturally given the correct unsupervised learning framework,
since they are generalizations that enable speech to be better predicted.
For these reasons, and also because there are many important engineering
problems where labeled training data is unavailable or expensive, we prefer
to investigate unsupervised methods.  A working unsupervised algorithm
would both dispel many of the learnability-argument myths surrounding child
language acquisition, and be a valuable tool to the natural language
engineering community.

\section{A Model of Speech Production}\label{speechmodel}

There is a conceptual difficulty with using a minimum description-length
learning framework when the input is a speech signal: speech is continuous,
and can be specified to an arbitrary degree of precision.  However, if we
assume that beyond a certain precision the variation is simply noise, it
makes sense to quantize the space of signals into small regions with fixed
volume $\Delta$. Given a probability density function $p(\cdot)$ over the
space of signals, the number of bits necessary to specify a region centered
at $u$ is approximately $-\log p(u)\Delta$.  The $\Delta$ contributes a
constant term that is irrelevant with respect to minimization, leaving the
effective description length of an utterance $u$ at $-\log p(u)$.  For
simplicity we follow much of the speech recognition community in computing
$p(u)$ from an intermediate representation, a sequence of {\em
  phones} $\phi$, using standard technology to compute $p(u|\phi)$.\footnote{In
  particular, phones are mapped to {\em triphones} by crossing them with
  features of their left and right context.  Each triphone is a 3-state HMM
  with Gaussian mixture models over a vector of mel-frequency cepstral
  coefficients and their first and second order differences.  Supervised
  training on the TIMIT data set is used for parameter estimation; this is
  explained in greater depth in section~\ref{speechhmms}.  Rabiner and
  Juang~\cite{Rabiner93} provide an excellent introduction to these
  methods.} More unusual, perhaps, is our model of the generation of this
phone sequence $\phi$, which attempts to capture some very rudimentary
aspects of phonology and phonetics.

\begin{figure}
\pageline
\begin{center}
\mbox{\ }
\psfig{figure=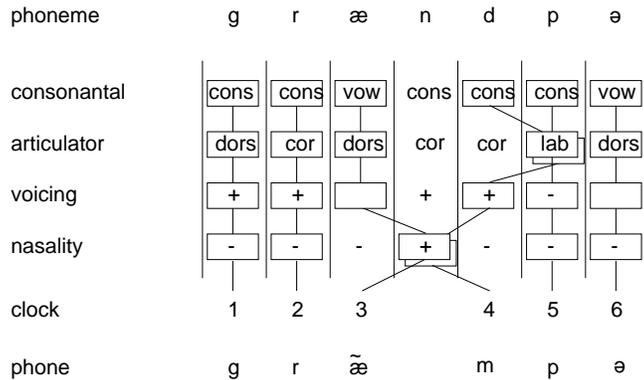,height=2in}
\end{center}

\caption{\label{fig:features}A 4-feature depiction of how
  /\unig\unir\uniA\unin\unid\unip\uniAX/ might surface as
  /\unig\unir\~{\uniA}\unim\unip\uniAX/.  The nasalization of the /\uniA/
  and /\unid/ and the place-of-articulation assimilation of the /d/ are
  explained by skew in the signals to those two articulators.  No phone is
  output for the phoneme /n/.} \pageline
\end{figure}

We adopt a natural and convenient model of the underlying representation of
sound in memory.  Each word is a sequence of feature bundles, or {\em
  phonemes}\footnote{A {\em phoneme} is a unit used to store articulatory
  information about a word in memory; a {\em phone} is similar but
  represents the commands actually sent to the articulation mechanism.  In
  our model, a sequence of phonemes undergoes various phonological
  transformations to become a sequence of phones; in general the set of
  phones is larger than the set of phonemes, though we constrain both to
  the set described in appendix~A.} (following Halle~\cite{Halle83b} these
features are taken to represent control signals to vocal articulators).
The fact that features are bundled is taken to mean that, in an ideal
situation, there will be some period when all articulators are in their
specified configurations simultaneously (this interpretation of
autosegmental representations is also taken by Bird and Ellison
\cite{Bird94}).  In order to allow for common phonological processes, and
in particular for the changes that occur in casual speech, we admit several
sources of variation between the underlying phonemes and the phones that
generates the speech signal.  In particular, a given phoneme may map to
zero, one, or more phones; features may exhibit up to one phoneme of skew
(this helps explain assimilation); and there is some inherent noise in the
mapping of features from underlying to surface forms.  Taken together, this
model can account for almost arbitrary insertion, substitution, and
deletion of articulatory features between the underlying phonemes and the
phone sequence, but it strongly favors changes that are expected given the
physical nature of the speech production process.
Figure~\ref{fig:features} contains a graphical depiction of a pronunciation
of the word {\em grandpa}, in which the underlying
/\unig\unir\uniA\unin\unid\unip\uniAX/ surfaces as
/\unig\unir\~{\uniA}\unim\unip\uniAX/.

To be more concrete, the underlying form of a sentence is taken to be the
concatenation of the phonemes of each word in the sentence.  This sequence
is mapped to the phone sequence by means of a stochastic finite-state
transducer, though of a much simpler sort than Kaplan and Kay use to model
morphology and phonology in their classic work~\cite{Kaplan94}: it has only
three states.  Possible actions are to {\em copy} (write a phone related to
the underlying phoneme and advance), {\em delete} (advance), {\em map}
(write a phone related to the underlying phoneme without advancing), and
{\em insert} (write an arbitrary phone without advancing).  When writing a
phone, the distribution over phones is a function of the features of the
current input phoneme, the next input phoneme, and the most recently
written phone.  The probability of deleting is related to the probability
of writing the same phone twice in succession.
Figure~\ref{fig:transitions} presents the state transition model for this
transducer; in our experiments the parameters of this model are fixed,
though obviously they could be re-estimated in later stages of learning.  In
the figure, $p_C(s|q,u,n)$ is the probability of the phone $s$ surfacing
given that the underlying phoneme $u$ is being copied in the context of the
previous phone $q$ and the subsequent phoneme $n$.  Similarly, $p_M(s|q,u)$
is the probability of mapping to $s$, and $p_I(s)$ is the probability of
inserting $s$.  Appendix~A describes how these functions are computed.

\begin{figure}
\pageline
\begin{small}
\begin{center}
\begin{tabular}{l|lll}
state & actions & next state & probability \\ \hline
{\em start}
& {\em insert}: $\mbox{write}(s)$ & {\em inserted} & $c_I\cdot p_I(s)$ \\

& {\em map}: $\mbox{write}(s)$ & {\em mapped} & $c_M\overline{c_I}
\cdot p_M(s|q,u)$ \\

& {\em delete}: $\mbox{advance}$ & {\em start} &
$\overline{c_M c_I}\cdot\min(c_D,p_C(q|q,u,n)+p_C(n|q,u,n))$ \\

& {\em copy}: $\mbox{advance}, \mbox{write}(s)$ & {\em start} &
$\overline{c_M c_I \cdot\min(\ldots)}\cdot p_C(s|q,u,n)$ \\
\\
{\em inserted}
& {\em map}: $\mbox{write}(s)$ & {\em mapped} & $c_M\cdot p_M(s|q,u)$ \\

& {\em delete}: $\mbox{advance}$ & {\em start} &
$\overline{c_M}\cdot\min(c_D,p_C(q|q,u,n)+p_C(n|q,u,n))$ \\

& {\em copy}: $\mbox{advance}, \mbox{write}(s)$ & {\em start} &
$\overline{c_M\cdot\min(\ldots)}\cdot p_C(s|q,u,n)$ \\
\\
{\em mapped}
& {\em map}: $\mbox{write}(s)$ & {\em mapped} & $c_M\cdot p_M(s|q,u)$ \\

& {\em copy}: $\mbox{advance}, \mbox{write}(s)$ & {\em start} &
$\overline{c_M}\cdot p_C(s|q,u,n)$

\end{tabular}
\end{center}
\end{small}

\caption{\label{fig:transitions}The state transitions of the transducer
    that maps from an underlying phoneme sequence to a sequence of
    phones.  Here, $s$ is the surface phone, $q$ is the previous surface
    phone, $u$ is the underlying phoneme, and $n$ is the next underlying
    phoneme.  In our current implementation, $c_I = 0.05, c_M = 0.05$ and
    $c_D = 0.9$ (chosen quite arbitrarily).}

\pageline
\end{figure}

This model of speech production determines $p(\phi|\pi)$, the probability
of a phone sequence $\phi$ given a phoneme sequence $\pi$, computed by
summing over all possible derivations of $\phi$ from $\pi$.  The
probability of an utterance $u$ given a phoneme sequence $\pi$ is then
$\sum_{\phi}p(u|\phi)p(\phi|\pi)$, and the combined description length of a
grammar $G$ and a set of utterances becomes $|G| + \sum_u -\log \sum_{\pi}
p(u|\pi)p_G(\pi)$.  Using this formula, any language model that assigns
probabilities to phoneme sequences can be evaluated with respect to the MDL
principle.  The problem of predicting speech is reduced to predicting
phoneme sequences; the extra steps do no more than contribute a term that
weighs phoneme sequences by how well they predict certain utterances.
Phonemes can be viewed as arbitrary symbols for the purposes of language
modeling, no different than text characters.  As a consequence, any
algorithm that can build a lexicon by modeling unsegmented text is a good
part of the way towards learning words from speech; the principal
difference is that in the case of speech there are two hidden layers that
must be summed over, namely the phoneme and phone sequences.  Some
approximations to this summation are discussed in sections~\ref{estimation}
and \ref{speechhmms}.  Meanwhile, the next two sections will treat the
input as a simple character sequence.

\section{The Class of Grammars}\label{grammar}

All unsupervised learning techniques rely on finding regularities in data.
In language, regularities exist at many different scales, from common
sound sequences that are words, to intricate patterns of grammatical
categories constrained by syntax, to the distribution of actions and
objects unique to a conversation.  These all interact to create weaker
second-order regularities, such as the high probability of the word {\em
  the} after {\em of}.  It can be extremely difficult to separate the
regularities tied to ``interesting'' aspects of language from those that
naturally arise when many complex processes interact.  For example, the
19-character sequence {\em radiopasteurization} appears six times in the
Brown corpus, far too often to be a freak coincidence.  But at the same
time, the 19-character sequence {\em scratching her nose} also appears
exactly six times.  Our intuition is that {\em radiopasteurization} is some
more fundamental concept, but it is not easy to imagine principled schemes
of determining this from the text alone.  The enormous number of
uninteresting coincidences in everyday language is distracting; plainly, a
useful algorithm must be capable of extracting fundamental regularities
even when such coincidences abound.  This and the minimum description
length principle are the motivation for our lexical representation (our
class of grammars).

In this class of grammars, terminals are drawn from an arbitrary alphabet.
For the time being, let us assume they are ascii characters, though in the
case of speech processing they are phonemes.  Nonterminals are concatenated
sequences of terminals.  Together, terminals and nonterminals are called
``words''.  The purpose of a nonterminal is to capture a statistical
pattern that is not otherwise predicted by the grammar.\footnote{Just as in
  the work of Cartwright and Brent~\cite{Cartwright94}, Della Pietra {\it
    et al}~\cite{DellaPietra95}, Olivier~\cite{Olivier68}, and
  Wolff~\cite{Wolff80,Wolff82}.} In this work, these patterns are merely
unusually common sequences of characters, though given a richer set of
linguistic primitives, the framework extends naturally.  As a general
principle, it is advantageous to add a word to the grammar when its
characters appear more often than can be explained given other knowledge,
though the cost of actually representing the word acts as a buffer against
words that occur only marginally more often than expected, or that have
unlikely (long) descriptions.

Some of the coincidences in the input data are of interest, and others are
not.  We assume that the vast majority of the less interesting coincidences
({\em scratching her nose}) arise from interactions between more
fundamental processes (verbs take noun-phrase arguments; {\em nose} is a
noun, and so on).  This suggests that fundamental processes can be
extracted by looking for patterns within the uninteresting coincidences,
and implies a recursive learning scheme: extract patterns from the input
(creating words), and extract patterns from those words, and so on.
These steps are equivalent to compressing not only the input, but also the
parameters of the compression algorithm, in a never-ending attempt to
identify and eliminate the predictable.  They lead us to a class of grammars
in which both the input and nonterminals are represented in terms of words.

\begin{figure}[t]
\pageline
\begin{small}
\begin{center}
\begin{tabular}{llcccr}
$\mbox{surf}(w)$ & $\mbox{rep}(w)$ & $c(w)$ & $p(w)$ & $-\log p(w)$ & $|\mbox{rep}(w)|$\\ \hline
the    & t+h+e   & 2 & 2/17 & 3.09 & 10.27\\
at     & a+t     & 2 & 2/17 & 3.09 &  6.18\\
t & & 2 & 2/17 & 3.09 & \\
h & & 2 & 2/17 & 3.09 & \\
cat    & c+at    & 1 & 1/17 & 4.09 &  7.18\\
hat    & h+at    & 1 & 1/17 & 4.09 &  6.18\\
thecat & the+cat & 1 & 1/17 & 4.09 &  7.18\\
thehat & the+hat & 1 & 1/17 & 4.09 &  7.18\\
e & & 1 & 1/17 & 4.09 & \\
a & & 1 & 1/17 & 4.09 & \\
c & & 1 & 1/17 & 4.09 & \\
i & & 1 & 1/17 & 4.09 & \\
n & & 1 & 1/17 & 4.09 & \\
\multicolumn{5}{l}{thecat+i+n+thehat} & 16.36\\ \cline{6-6}
& & & & & 60.53
\end{tabular}
\end{center}
\end{small}
\vspace{-.15in}
\caption{\label{fig:model}A possible description of {\em
    thecatinthehat}, with length 60.53.}
\pageline
\end{figure}

Given that our only unit of representation is the word, compression of the
input or a nonterminal reduces to writing out a sequence of word indices.
For simplicity, these words are drawn from a probability distribution over a
single dictionary; this language model has been called a {\em
  multigram}~\cite{Deligne95}.  Figure~\ref{fig:model} presents a complete
description of {\em thecatinthehat}, in which the input and six words used
in the description are decomposed using a multigram language model.  This
is a contrived example, and does not represent how our algorithm would
analyze this input.  The input is represented by four words,
{\em thecat+i+n+thehat}.  The surface form of a word $w$ is given by
$\mbox{surf}(w)$, and its representation by $\mbox{rep}(w)$.  The total
number of times the word is indexed in the combined description of the
input and the dictionary is $c(w)$.  Using maximum-likelihood estimation,
the probability $p(w)$ of a word is computed by normalizing these counts.
Assuming a clever coding, the length of a word $w$'s index is $-\log p(w)$.
The cost of representing a word in the dictionary is the total length of
the indices in its representation; this is denoted by $|\mbox{rep}(w)|$
above.\footnote{Of course, the number of words in the representation must
  also be written down, but this is almost always negligible compared to
  the cost of the indices.  For that reason, we do not discuss it further.
  The careful reader will notice an even more glaring omission-
  nowhere are word indices paired with words.  However, if words are
  written in rank order by probability, they can be uniquely paired with
  Huffman codes given only knowledge of how many codes of each length
  exist.  For large grammars, the length of this additional information is
  negligible.  We have found that Huffman codes closely approach the
  arithmetic-coding ideal for this application.}
(Terminals have no representation or cost.)  The total description length
is this example is $60.53$ bits, the summed lengths of the representations
of the input and all the nonterminals. This is longer than the
``empty'' description, containing no nonterminals: for this short input the
words we contrived were not justified.  But because only $16.36$ bits of
the description were devoted to the input, doubling it to {\em
  thecatinthehatthecatinthehat} would add no more than $16.36$ bits to the
description length, whereas under the empty grammar the description length
doubles.  That is because in this longer input, the sequences {\em thecat}
and {\em thehat} appear more often than independent chance would predict,
and are more succinctly represented by a single index than by writing down
their letters piecemeal.

This representation is a generalization of that used by the
LZ78~\cite{Ziv78} coding scheme.  It is therefore capable of universal
compression, given the right estimation scheme, and compresses a sequence
of identical characters of length $n$ to size $O(\log n)$.  It has a
variety of other pleasing properties.  Because each word appears in the
dictionary only once, common idiomatic or suppletive forms do not unduly
distort the overall picture of what the ``real'' regularities are, and the
fact that commonly occuring patterns are compiled out into words also
explains how a phrase like /\uniw\uniAH\uniC\uniAX\unid\uniu\uniIX\unin/
can be recognized so easily.

\section{Finding the Optimal Grammar}\label{search}

The language model in figure~\ref{fig:model} looks suspiciously like a
stochastic context-free grammar, in that a word is a nonterminal that
expands into other words.  Context-free grammars, stochastic or not, are
notoriously difficult to learn using unsupervised algorithms.  As a general
rule, CFGs acquired this way have neither achieved the entropy rate of
theoretically inferior Markov and hidden Markov processes \cite{Chen95},
nor settled on grammars that accord with linguistic intuitions
\cite{Carroll92,Pereira92} (for a detailed explanation of why this is so,
see de Marcken \cite{deMarcken95b}).  However disappointing previous
results have been, there is reason to be optimistic.  First of all, as
described so far the class of grammars we are considering is weaker than
context-free: there is recursion in the language that grammars are
described with, but not in the languages these grammars generate.  In fact,
the grammars make no use of context whatsoever.  Expressive power has not
been the principle downfall of CFG induction schemes, however: the search
space of stochastic CFGs under most learning strategies is riddled with
local optima.  This means that convergence to a global optimum using a
hill-climbing approach like the inside-outside algorithm \cite{Baker79} is
only possible given a good starting point, and there are arguments
\cite{Carroll92,deMarcken95b} that algorithms will not usually start from
such points.

Fortunately, the form of our grammar permits the use of a significantly
better behaved search algorithm.  There are several reasons for this.
First, because each word is decomposable into its representation, adding or
deleting a word does not drastically alter the character of the grammar.
Second, because all of the information about a word necessary for parsing
is contained in its surface form and its probability, its representation is
free to change abruptly from one iteration to the next, and is not tied to
the history of the search process.  Finally, because the representation of
a word serves as a prior that discriminates against unnatural words, search
tends not to get bogged down in linguistically implausible grammars.

The search algorithm we use is divided into four stages.  In stage~1, the
Baum-Welch~\cite{Baum70} procedure is applied to the input and word
representations to estimate the probabilities of the words in the current
dictionary.  In stage~2 new words are added to the dictionary if this is
predicted to reduce the combined description length of the dictionary and
input.  Stage~3 is identical to stage~1, and in stage 4 words are deleted
from the dictionary if this is predicted to reduce the combined description
length.  Stages~2 and 4 are a means of increasing the likelihood of the
input by modifying the contents of the dictionary rather than the
probability distribution over it, and are thus part of the maximization
step of a generalized expectation-maximization (EM)
procedure~\cite{Dempster77}.  Starting from a dictionary that contains only
the terminals, these four stages are iterated until the dictionary
converges.

\subsection{Probability Estimation}\label{estimation}

Given a dictionary, we can compute word probabilities over word and input
representations using EM; for the language model described here this is
simply the Baum-Welch procedure.  For a sequence of terminals
$\subseq{0}{t}{n}$ (text characters or phonemes), and a word $w$ that can
span a portion of the sequence from $k$ to $l$ ($\mbox{surf}(w) =
\subseq{k}{t}{l}$), we write $\kwl$.  Since there is only one state in the
hidden Markov formulation of multigrams (word emission is
context-independent), the form of the algorithm is quite simple:

\vspace{-.3in}
\begin{eqnarray*}
\alpha_{k} & \equiv & p(\subseq{0}{t}{k}) = \wordsumlwk p(w)\alpha_l.\\
\beta_{k} & \equiv & p(\subseq{k}{t}{n}) = \wordsumkwl p(w)\beta_l.\\
p(l\stackrel{w}{\rightarrow}k|\subseq{0}{t}{n})
& \equiv & \frac{p(\subseq{0}{t}{n},\lwk)}{p(\subseq{0}{t}{n})} =\frac{\alpha_l p(w)\beta_k}{\beta_0}.
\end{eqnarray*}
\vspace{-.3in}

\noindent with $\alpha_0 = \beta_n = 1$.  Summing the posterior probability
$p(\lwk|\subseq{0}{t}{n})$ of a word $w$ over all possible locations
produces the expected number of times $w$ is used in the combined
description.  Normalizing these counts produces the next round of word
probabilities.  These two steps are iterated until convergence; two or
three iterations usually suffice.  The above equations are for
complete-likelihood estimates, but if one adopts the philosophy that a word
has only one representation, a Viterbi formulation can be used.  We have
not noticed that the choice leads to significantly different results in
practice\footnote{All results in this paper except for tests on raw speech
  use the complete-likelihood formulation.}, and a Viterbi implementation
can be simpler and more efficient.

There are two complications that arise in the estimation.  The first is
quite interesting.  For a description to be well-defined, the
graph of word representations can not contain cycles: a word can not be
defined in terms of itself.  So some partial ordering must be imposed on
words.  Under the concatenative model that has been discussed, this is easy
enough, since the representation of a word can only contain shorter words.  But
there are obvious and useful extensions that we have experimented with,
such as applying the phoneme-to-phone model at every level of
representation, so that a word like {\em wanna} can be represented in terms
of {\em want} and {\em to}.  In this case, a chicken-and-egg problem must
be solved: given two words, which comes first?  It is not easy to find good
heuristics for this problem, and computing the description length of all
possible orderings is obviously far too expensive.

The second problem is that when the forward-backward algorithm is extended
with the bells-and-whistles necessary to accommodate the phoneme-to-phone
model (even ignoring the phone-to-speech layer), it becomes quite expensive, for
three reasons.  First, a word can with some probability match anywhere in
an utterance, not just where characters align perfectly.  Second, the
position and state of the read head for the phone-to-phoneme model must be
incorporated into the state space.  And third, the probability of the
word-to-phone mapping is now dependent on the first phoneme of the next
word.  Without going into the lengthy details, our algorithm is made
practical by prioritizing states for expansion by an estimate of their
posterior probability, computed from forward ($\alpha$) and backward
($\beta$) probabilities.  The algorithm interleaves state expansion with
recomputation of forward and backward probabilities as word matches are
hypothesized, and prunes states that have low posterior estimates.  Even
so, running the algorithm on speech is two orders of magnitude slower than
on text.

\subsection{Adding and Deleting Words}

The governing motive for changing the dictionary is to reduce the combined
description length of $U$ and $G$, so any improvement a new word brings to
the description of an utterance must be weighed against its representation
cost.  The general strategy for building new words is to look for a set of
existing words that occur together more often than independent chance would
predict.\footnote{In our implementation, we consider only sequences of two or
  three words that occurs in the Viterbi analyses of the
  word and input representations.} The addition of a new word with the same
surface form as this set will reduce the description length of the
utterances it occurs in.  If its own cost is less than the reduction, the
word is added.  Similarly, words are deleted when doing so would
reduce the combined description length.  This generally occurs as shorter
words are rendered irrelevant by longer words that model more of the input.

Unfortunately, the addition or deletion of a word from the grammar could
have a substantial and complex impact on the probability distribution
$p(w)$.  Because of this, it is not possible to efficiently gauge the exact
effect of such an action on the overall description length, and various
approximations are necessary.  Rather than devote space to them here, they
are described in appendix~B, along with other details related to the
addition and deletion of words from the dictionary.

One interesting addition needed for processing speech is the ability to
merge changes that occur in the phoneme-to-phone mapping into existing
words.  Often, a word is used to match part of a word with different
sounds; for instance {\em doing} /\unid\uniu\uniI\uniG/ may initially be
analyzed as {\em do} /\unid\uniu/ + {\em in} /\uniI\unin/, because {\em in}
is much more probable than {\em -ing}.  This is a common pair that will be
joined into a single new word.  Since in most uses of this word the /\unin/
changes to a /\uniG/, it is to the algorithm's advantage to notice this and
create /\unid\uniu\uniI\uniG/ from /\unid\uniu\uniI\unin/.  The other
possible approach, to build words based on the surface forms found in the
input rather than the concatenation of existing words, is less attractive,
both because it is computationally more difficult to estimate the effect of
adding such words, and because surface forms are so variable.

\section{Experiments and Results}\label{results}

There are at least three qualities we hope for in our algorithm.  The first
is that it captures regularities in the input, using as efficient a model
as possible.  This is tested by its performance as a text-compression and
language modeling device.  The second is that it captures regularity
using linguistically meaningful representations.  This is tested by using
it to compress unsegmented phonetic transcriptions and then verifying that
its internal representation follows word boundaries.  Finally, we wish it
to learn given even the most complex of inputs.  This is tested by applying
the algorithm to a multi-speaker corpus of continuous speech.

\subsection{Text Compression and Language Modeling}

The algorithm was run on the Brown corpus~\cite{Francis82}, a collection of
approximately one million words of text drawn from diverse sources, and a
standard test of language models.  We performed a test identical to Ristad
and Thomas~\cite{Ristad95}, training on 90\% of the corpus\footnote{The
  first 90\% of each of the 500 files.  The alphabet is 70 ascii
  characters, as no case distinction is made.  The input was segmented into
  sentences by breaking it at all periods, question marks and exclamation
  points.  Compression numbers include all bits necessary to exactly
  reconstruct the input.} and testing on the remainder.  Obviously, the
speech extensions discussed in section~\ref{speechmodel} were not
exercised.

After fifteen iterations, the training text is compressed from 43,337,280
to 11,483,361 bits, a ratio of 3.77:1 with a compression rate of 2.12
bits/character; this compares very favorably with the 2.95 bits/character
achieved by the LZ77 based {\em gzip} program.  9.5\% of the description is
devoted to the parameters (the words and other overhead), and the rest
to the text.  The final dictionary contains 30,347 words.  The
entropy rate of the training text, omitting the dictionary and other
overhead, is 1.92 bits/character.  The entropy rate of this same language
model on the held-out test set (the remaining 10\% of the corpus) is 2.04
bits/character.  A slight adjustment of the conditions for creating words
produces a larger dictionary, of 42,668 words, that has a slightly poorer
compression rate of 2.19 bits/character but an entropy rate on the test
set of 1.97 bits/character, identical to the base rate Ristad and Thomas
achieve using a non-monotonic context model.  So far as we are aware, all
models that better our entropy figures contain far more parameters and do
not fare well on the total description-length (compression rate)
criterion.  This is impressive, considering that the simple language model
used in this work has no access to context, and naively reproduces
syntactic and morphological regularities time and time again for words with
similar behavior.

\begin{figure}
  \pageline
\begin{small}
$|$ ordinary$|$ care$|$y$|$ williams$|$,$|$ armed with$|$ a$|$ pistol$|$,$|$ stood$|$ by$|$ at the$|$
$|$ poll$|$s$|$ to insure$|$ order$|$.$|$    "this$|$ was$|$ the$|$ cool$|$est$|$,$|$ calm$|$est$|$
$|$ election$|$ i ever saw$|$\",$|$ col$|$qui$|$tt$|$ policeman$|$ to$|$m$|$ williams$|$ said.$|$
$|$ "be$|$ing$|$ at the$|$ poll$|$s$|$ was$|$ just like$|$ being$|$ at$|$ church$|$.$|$ i didn't$|$
$|$ smell$|$ a$|$ drop$|$ of$|$ liquor$|$, and we$|$ didn't have$|$ a bit of$|$ trouble$|$
$|$".$|$    the$|$ campaign$|$ leading$|$ to the$|$ election$|$ was not$|$ so$|$ quiet$|$
$|$, however$|$.$|$ it was$|$ marked$|$ by$|$ controversy$|$,$|$ anonymous$|$ midnight$|$ phone$|$
$|$ calls$|$ and$|$ veiled$|$ threats$|$ of$|$ violence$|$.$|$   $|$ the former$|$ county$|$
$|$ school superintendent$|$,$|$ george$|$ p\&$|$ call$|$an$|$,$|$ shot$|$ himself$|$ to death$|$
$|$ march$|$ 18$|$,$|$ four$|$ days after$|$ he$|$ resigned$|$ his$|$ post$|$ in a$|$ dispute$|$
$|$ with the$|$ county$|$ school board$|$.$|$   $|$ during the$|$ election$|$ campaign$|$,$|$
$|$ both$|$ candidates$|$,$|$ davis$|$ and$|$ bush$|$,$|$ reportedly$|$ received$|$ anonymous$|$
$|$ telephone calls$|$.$|$ ordinary$|$ williams$|$ said he$|$, too,$|$ was$|$ subjected to$|$
$|$ anonymous$|$ calls$|$ soon$|$ after$|$ he$|$ scheduled$|$ the$|$ election$|$.$|$   $|$
$|$ many$|$ local$|$ citizens$|$ feared$|$ that$|$ there would be$|$ irregularities$|$
$|$ at the$|$ poll$|$s$|$, and$|$ williams$|$ got$|$ himself$|$ a$|$ permit$|$ to carry$|$ a$|$
$|$ gun$|$ and$|$ promised$|$ an$|$ orderly$|$ election$|$.$|$

\end{small}
\caption{\label{fig:text}The top-level segmentation of the first 10 sentences of the
  test set, previously unseen by the algorithm.  Vertical bars indicate
  word boundaries.}
\pageline
\end{figure}

\begin{figure}
  \pageline \vspace{-.15in}
\begin{small}
\begin{center}
\begin{tabular}{rrrrll}
Rank & $-\log p(w)$ & $|\mbox{rep}(w)|$ & $c(w)$ & $\mbox{rep}(w)$\\ \hline 
    0 &  4.653 &        & 38236.34 & \verb*|.| \\
    1 &  4.908 &        & 32037.72 & \verb*|,| \\
    2 &  5.561 & 20.995 & 20369.20 & \verb*|[ [the]]| \\
    3 &  5.676 &        & 18805.20 & \verb*|s| \\
    4 &  5.756 & 17.521 & 17791.95 & \verb*|[[ an]d]| \\
    5 &  6.414 & 22.821 & 11280.19 & \verb*|[ [of]]| \\
    6 &  6.646 & 18.219 & 9602.04 & \verb*|[ a]| \\
  100 & 10.316 &        &  754.54 & \verb*|o| \\
  101 & 10.332 & 20.879 &  746.32 & \verb*|[ [me]]| \\
  102 & 10.353 & 24.284 &  735.25 & \verb*|[ [two]]| \\
  103 & 10.369 & 21.843 &  727.30 & \verb*|[ [time]]| \\
  104 & 10.379 & 23.672 &  722.46 & \verb*|[ (]| \\
  105 & 10.392 & 18.801 &  715.74 & \verb*|["?]| \\
  106 & 10.434 &        &  694.99 & \verb*|m| \\
  500 & 12.400 & 19.727 &  177.96 & \verb*|[ce]| \\
  501 & 12.400 &        &  177.94 & \verb*|2| \\
  502 & 12.401 & 21.364 &  177.86 & \verb*|[[ize]d]| \\
  503 & 12.402 & 16.288 &  177.72 & \verb*|[[   ][ but]]| \\
  504 & 12.403 & 21.053 &  177.60 & \verb*|[ [con]]| \\
  505 & 12.408 & 21.346 &  176.94 & \verb*|[[ to][ok]]| \\
  506 & 12.410 & 24.251 &  176.74 & \verb*|[ [making]]| \\
 1000 & 13.141 & 22.861 &  106.50 & \verb*|[[ require]d]| \\
 1001 & 13.141 & 22.626 &  106.49 & \verb*|[i[ous]]| \\
 1002 & 13.142 & 17.065 &  106.39 & \verb*|[[ed by][ the]]| \\
 1003 & 13.144 & 29.340 &  106.24 & \verb*|[[ ear][lier]]| \\
 1004 & 13.148 & 24.391 &  105.96 & \verb*|[ [paid]]| \\
 1005 & 13.148 & 20.041 &  105.94 & \verb*|[be]| \\
 1006 & 13.149 & 21.355 &  105.89 & \verb*|[[ clear][ly]]| \\
28241 & 18.290 & 33.205 &    3.00 & \verb*|[[ massachusetts][ institute of technology]]| \\
30000 & 18.875 & 60.251 &    2.00 & \verb*|[[ norman][ vincent][ pea][le]]|\\
30001 & 18.875 & 61.002 &    2.00 & \verb*|[[ pi][dding][ton][ and][ min]]|\\
30002 & 18.875 & 69.897 &    2.00 & \verb*|[[ **f where the maximization is][ over]| \\
      &        &        &         & \verb| |\verb*|[ all][ admissible][ policies]]| \\
30003 & 18.875 & 69.470 &    2.00 & \verb*|[[ stick to][ an un][charg][ed][ surface]]| \\
30004 & 18.875 & 63.360 &    2.00 & \verb*|[[ mother][-of-]p[ear]l]| \\
30005 & 18.875 & 61.726 &    2.00 & \verb*|[[ gov&][ mar][vin][ griffin]]| \\
30006 & 18.875 & 55.739 &    2.00 & \verb*|[[ reacted][ differently][ than][ they had]]| \\
\end{tabular}
\end{center}
\end{small}

\caption{\label{fig:words}Some words from the dictionary with their representations, ranked by probability.}
\pageline
\end{figure}

Figure~\ref{fig:text} presents the segmentation of the first 10 sentences
of the held-out test set, and figure~\ref{fig:words} presents a subset of
the final dictionary.  Even with no special knowledge of the space
character, the algorithm adopts a policy of placing spaces at the start of
words.  The words in figure~\ref{fig:words} with rank 1002 and 30003 are
illuminating: they are cases where syntactic and morphological regularities
are sufficient to break word boundaries; solutions to this ``problem'' are
discussed in section~\ref{extensions}.  Many of the rarer words are
uninteresting coincidences, useful for compression only because of
the peculiarities of the source.\footnote{The author apologizes for the
  presumably inadvertent addition of word 28241 to the sample.}

\subsection{Segmentation}

\begin{figure}
\pageline
\begin{small}
\begin{tabular}{l}
this is a book ?\\
what do you see in the book ?\\
how many rabbits ?\\
how many ?\\
one rabbit ?\\ \cline{1-1}
uhoh trouble what else did you forget at school ?\\
we better go on Monday and pick up your picture and your dolly .\\
would you like to go to that school ?\\
there 're many nice people there weren't there ?\\
did they play music ?\\ \cline{1-1}
oh what shall we do at the park ?\\
oh good good good .\\
I love horses .\\
here we go trot trot trot trot .\\
and now we 'll go on a merry go round .
\end{tabular}
\end{small}
\caption{\label{fig:childes} Several short excerpts from the Nina portion
  of the CHILDES database.}
\pageline
\end{figure}

The algorithm was run on a collection of 34,438 transcribed sentences of
mothers' speech to children, taken from the Nina portion of the CHILDES
database~\cite{MacWhinney85}; a sample is shown in
figure~\ref{fig:childes}.  These sentences were run through a simple
public-domain text-to-phoneme converter, and inter-word pauses were
removed.  This is the same input described in de
Marcken~\cite{deMarcken94b}.  Again, the phoneme-to-phone portion of our
work was not exercised; the output of the text-to-phoneme converter is free
of noise and makes this problem little different from that of segmenting
text with the spaces removed.

The goal, as in Cartright and Brent~\cite{Cartwright94}, is to segment the
speech into words.  After ten iterations of training on the phoneme
sequences, the algorithm produces a dictionary of 6,630 words, and a
segmentation of the input.  Because the text-to-phoneme engine we use is
particularly crude, each word in the original text is mapped to a
non-overlapping region of the phonemic input.  Call these regions the {\em
  true} segmentation.  Then the {\em recall rate} of our algorithm is
96.2\%: fully 96.2\% of the regions in the true segmentation are exactly spanned
by a single word at {\em some} level of our program's hierarchical
segmentation.  Furthermore, the {\em crossing rate} is 0.9\%: only 0.9\% of
the true regions are partially spanned by a word that also spans some
phonemes from another true region.  Performing the same evaluations
after training on the first 30,000 sentences and testing on the remaining
4,438 sentences produces a recall rate of 95.5\% and a crossing rate of
0.7\%.

Although this is not a difficult task, compared to that of segmenting raw
speech, these figures are encouraging.  They indicate that given simple
input, the program very reliably extracts the fundamental linguistic units.
Comparing to the only other similar results we know of, Cartwright and
Brent's~\cite{Cartwright94}, is difficult: at first glance our recall figure seems dramatically
better, but this is partially because of our multi-level representation,
which also renders accuracy rates meaningless.  We are not aware of
other reported crossing rate figures.

\subsection{Acquisition from Speech}\label{speechhmms}

The experiments we have performed on raw speech are preliminary, and
included here principally to demonstrate that our algorithm {\em does}
learn words even in the very worst of conditions.  The conditions of these
initial tests are so extreme to make detailed analysis irrelevant, but we
believe the final dictionaries are convincing in their own right.

A phone-to-speech model was created using {\em supervised}
training\footnote{The ethics of using supervised training for this portion
  of an otherwise unsupervised algorithm do not overly concern us.  We
  would prefer to use a phone-to-speech model based directly on
  acoustic-phonetics, and there is a wide literature on such methods, but
  the HMM recognizer is more convenient given the limited scope of this
  work.  It is unlikely that the substitution of a different model would
  reduce the performance of our algorithm, given the high error rate of the
  HMM recognizer.  We also test our algorithm on some of the same speech
  used for training the models, because there is a very limited amount of
  data available in the corpus.  Again, for these preliminary tests this
  should not be a great concern.} on the TIMIT continuous speech database.
Specifically, the HTK HMM toolkit developed by Young and Woodland was used
to train a triphone based model on the `si' and `sx' sentences of the
database.  Tests on the TIMIT test set put phone recall at 55.5\% and phone
accuracy at 68.7\%.  These numbers were computed by comparing the Viterbi
analyses of utterances under a uniform prior to phoneticians'
transcriptions.  It should be clear from this performance level that the
input to our algorithm will be very, very noisy: some sentences with their
transcriptions and the output of the phone recognizer are presented in
figure~\ref{fig:timit}.  Because of the extra computational expense
involved in summing over different phone possibilities only the Viterbi
analyses were used; errors in the Viterbi sequences must therefore be
compensated for by the phoneme-to-phone model. We intend to extend
the search to a network of phones in the near future.

\begin{figure}
\pageline
\begin{small}
\def\ipa{\ipatenrm}
\begin{tabular}{ll}
\multicolumn{2}{l}{Bricks are an alternative.} \\
\hspace{.15in} &
\unib\unir\uniI\unik\unis\unit\unia\unir\unin\uniO\unil\unit\unir\unin\uniIX\unit\uniI\univ
{\em (trans.) }\\
\hspace{.15in} &
\unib\unir\uniIX\unik\uniz\unia\unir\uniE\unin\uniO\unil\unit\unir\uniIX\unin\uniIX\unit\uniI\univ
{\em (rec.)} \\ \cline{1-2}
\multicolumn{2}{l}{Fat showed in loose rolls beneath the shirt.} \\
&
\unif\uniA\unit\uniS\unio\uniu\unid\unit\uniIX\unin\unil\uniu\unis\unir\unio\uniu\unil\uniz\unib\uniIX\unin\unii\uniT\uniIX\uniS\unir\unit
{\em (trans.) }\\
&
\unif\uniA\unit\uniS\uniE\unid\unii\uniI\unin\unid\uniAX\uniD\uniAX\unil\uniIX\unis\uniw\unir\unil\unit\unis\unip\uniIX\unit\unin\unii\uniIX\uniT\uniD\uniIX\uniS\unir\unit
{\em (rec.)} \\ \cline{1-2}
\multicolumn{2}{l}{It suffers from a lack of unity of purpose and respect for heroic leadership.} \\
&
\uniIX\unit\unis\uniAH\unif\unir\uniz\unif\unir\uniAX\unim\uniAX\unil\uniA\unik\uniIX\univ\uniy\uniu\unin\uniIX\unit\unii\uniAX\univ\unip\unir\unip\uniIX\unis\uniE\unin\unir\uniIX\unis\unip\uniE\unik\unit\unif\unir\uniH\unir\unio\uniu\uniIX\unik\unil\unii\unit\unir\uniS\uniI\unip
{\em (trans.) } \\
&
\uniIX\unit\unis\uniT\uniAH\unip\unir\uniz\unif\unir\unin\unia\unil\uniA\unik\uniE\unid\unik\unii\uniIX\unin\uniI\unid\unis\uniEPI\unii\uniIX\unip\unir\unip\uniAH\unis\uniIX\unin\unir\uniIX\unis\unip\unib\uniA\unik\unit\unif\unir\unih\unir\uniA\unil\uniIX\unik\unil\unii\unir\uniS\uniA\unip
{\em (rec.)} \\ \cline{1-2}
\multicolumn{2}{l}{His captain was thin and haggard and his beautiful boots were worn and shabby.} \\
&
\unih\uniI\uniz\unik\uniA\unip\unit\uniIX\unin\uniw\uniAX\unis\uniT\uniI\unin\uniA\unin\uniH\uniA\unig\unir\unid\uniIX\unin\uniI\uniz\unib\uniy\uniu\unit\uniu\unif\unil\unib\uniu\unit\unis\uniEPI\uniw\unir\uniw\uniO\unir\unin\uniIX\unin\uniEPI\uniS\uniA\unib\unii
{\em (trans.) } \\
&
\unih\uniI\uniz\unik\unia\unit\uniAH\unin\uniw\uniAX\unis\unit\uniD\uniA\unin\uniAX\unin\unih\uniA\unig\uniI\unir\unid\uniE\unin\uniI\uniIX\uniz\unip\unib\uniy\uniu\unit\uniIX\unif\unil\unid\unib\unil\unio\uniu\unik\unit\uniz\uniEPI\uniw\unir\uniw\uniO\unir\unin\uniIX\uniG\uniS\uniA\unib\unii
{\em (rec.)} \\ \cline{1-2}
\multicolumn{2}{l}{The reasons for this dive seemed foolish now.} \\
&
\uniD\uniIX\unir\unii\uniz\uniAX\unin\uniz\unif\unir\uniD\uniI\unis\unid\unia\unii\univ\unis\unii\unim\unid\unif\uniu\unil\uniIX\uniS\uniEPI\unin\unia\uniu
{\em (trans.) } \\
&
\uniD\uniIX\unir\unii\unis\uniA\unin\unis\unip\uniAX\unid\uniIX\unis\unit\unib\unia\unii\unie\unii\unib\unip\unis\unii\uniG\unig\unid\unif\unil\unio\uniu\unin\unil\uniIX\uniS\uniEPI\unin\unia\uniu\unil
{\em (rec.)}

\end{tabular}
\end{small}

\caption{\label{fig:timit} The first 5 of the 1890 sentences used to
  test our algorithm; both the transcriptions and the phone recognizer's
  output are shown.  TIMIT's phone set has been mapped into the set from
  appendix~A; in the process some information has been lost (for instance,
  the epenthetic vowels that often occur before syllabic consonants
  have been deleted).}
\pageline
\end{figure}

We ran the algorithm on the 1890 `si' sentences from TIMIT, both on the raw
speech using the Viterbi analyses and on the cleaner transcriptions.  This
is a very difficult training corpus: TIMIT was designed to aid in the
training of acoustic models for speech recognizers, and as a consequence
the source material was selected to maximize phonetic diversity.  Not
surprisingly, therefore, the source text is very irregular and contains few
repetitions.  It is also small.  As a final complication, the
sentences in the corpus were spoken by hundreds of different speakers of
both sexes, in many different dialects of English.  We hope in the near
future to apply the algorithm to a longer corpus of dictated Wall Street
Journal articles; this should be a fairer test of performance.

\begin{figure}
\pageline
\begin{small}
\def\ipa{\ipatenrm}
\begin{center}
\begin{tabular}{rllrll}
\multicolumn{3}{c}{Speech} & \multicolumn{3}{c}{Transcriptions} \\
Rank & $\mbox{rep}(w)$ & Interpretation & Rank & $\mbox{rep}(w)$ &
Interpretation \\ \hline
    0 & \unit & &
  100 & \verb|[|\uniD\uniE\unir\verb|]| & {\em there, their}\\
    1 & \unid & &
  101 & \verb|[|\uniw\uniI\verb|]| \\
    2 & \unis & &
  102 & \verb|[|\uniAH\verb|[|\uniD\unir\verb|]|\verb|]| & {\em  other} \\
    3 & \unik & &
  103 & \verb|[|\unis\uniAH\unim\verb|]| & {\em some}\\
   20 & \verb|[|\uniIX\uniz\verb|]| & {\em -es, is} &
  500 & \verb|[|\unip\verb|[|\uniO\unii\verb|]|\unin\unit\verb|]| & {\em point} \\
   21 & \verb|[|\uniI\unin\verb|]| & {\em in} &
  501 & \verb|[|\unik\uniI\unid\verb|]| \\
   22 & \verb|[|\unih\unii\verb|]| & {\em he} &
  502 & \verb|[|\uniIX\verb|[|\unib\unil\verb|]|\verb|]| & {\em -able} \\
   23 & \verb|[|\uniE\unin\verb|]| & &
  503 & \verb|[|\unis\verb|[|\unit\unie\unii\verb|]|\verb|]| & {\em stay}\\
   24 & \univ & &
  504 & \verb|[|\verb|[|\unid\uniJ\uniE\unin\unir\verb|]|\unil\verb|]| & {\em general}\\
   25 & \verb|[|\uniD\uniAX\verb|]| & {\em the} &
  740 & \verb|[|\verb|[|\unip\unir\unia\verb|]|\verb|[|\unib\unil\verb|]|\verb|]| \\
  250 & \verb|[|\verb|[|\unip\unir\verb|]|\unii\verb|]| & {\em pre-}&
  741 & \verb|[|\verb|[|\unia\unii\unid\unii\verb|]|\uniAX\uniz\unir\verb|]| \\
  251 & \verb|[|\verb|[|\uniw\uniE\verb|]|\unil\verb|]| & {\em well} &
  742 & \verb|[|\verb|[|\unif\uniI\unil\verb|]|\unim\verb|]| & {\em film} \\
  252 & \verb|[|\unid\uniI\unit\verb|]| & &
  743 & \verb|[|\uniI\unib\uniu\verb|]| \\
  253 & \verb|[|\unia\unis\unip\verb|]| & &
  744 & \verb|[|\unif\uniU\unit\verb|]| & {\em foot} \\
  254 & \verb|[|\uniAH\verb|[|\uniD\unir\verb|]|\verb|]| & {\em other} &
  745 & \verb|[|\unif\verb|[|\unil\unia\uniu\verb|]|\verb|]| & {\em flow(er)} \\
  310 & \verb|[|\verb|[|\unim\unie\unii\verb|]|\verb|[|\unib\unii\verb|]|\verb|]| & {\em maybe}&
  746 & \verb|[|\unil\verb|[|\uniA\unit\verb|]|\unir\verb|]| & {\em latter}\\
  311 & \verb|[|\verb|[|\uniw\unii\verb|]|\uniG\verb|]| & &
  747 & \verb|[|\verb|[|\unis\uniAH\unim\verb|]|\verb|[|\unit\unia\unii\unim\verb|]|\verb|]| & {\em sometime}\\
  312 & \verb|[|\unin\verb|[|\uniE\univ\unir\verb|]|\verb|]| & {\em never} &
  748 & \verb|[|\uniAX\unil\unia\verb|[|\unid\uniJ\verb|]|\verb|[|\uniIX\unik\unil\verb|]|\verb|]| & {\em -ological} \\
  313 & \verb|[|\verb|[|\unia\unii\verb|]|\uniIX\unit] & &
  749 & \verb|[|\verb|[|\unik\uniw\uniE\verb|]|\uniS\verb|[|\unit\uniC\verb|]|\verb|]| & {\em quest(ion)}\\
  314 & \verb|[|\unii\verb|[|\unit\uniC\verb|]|\verb|]| & {\em each}&
 1070 & \verb|[|\verb|[|\uniAH\unin\verb|]|\verb|[|\unif\uniO\unir\verb|]|\verb|[|\unit\uniC\verb|]|\verb|]| & {\em unfort(unate)}\\
  315 & \verb|[|\unil\verb|[|\uniE\unit\verb|]|\verb|]| & {\em let}&
 1071 & \verb|[|\verb|[|\unit\uniIX\verb|]|\unip\uniAX\verb|[|\unik\unil\verb|]|\verb|]| & {\em typical}\\
  714 & \verb|[|\verb|[|\uniw\uniU\verb|]|\unit\verb|[|\uniE\univ\unir\verb|]|\verb|]| & {\em whatever}&
 1072 & \verb|[|\unit\verb|[|\uniI\unis\unit\verb|]|\verb|[|\uniw\uniIX\uniz\verb|]|\verb|]| \\
  715 & \verb|[|\verb|[|\uniD\uniIX\verb|]|\verb|[|\unit\uniC\uniI\unil\verb|]|\unid\verb|]| &{\em the child(ren)} &
 1073 & \verb|[|\verb|[|\unis\uniE\unit\verb|]|\verb|[|\uniAX\unis\unia\unii\unid\unit\verb|]|\verb|]| & {\em set aside}\\
  716 & \verb|[|\unis\uniI\unis\verb|[|\uniAX\unim\verb|]|\verb|]| & {\em system}&
 1074 & \verb|[|\verb|[|\unir\unii\verb|]|\uniz\verb|[|\uniAH\unil\verb|]|\verb|]| & {\em resul(t)}\\
  717 &  \verb|[|\verb|[|\unik\uniAH\unim\verb|]|\unip\verb|[|\uniD\uniAX\verb|]|\verb|]| & &
 1075 & \verb|[|\verb|[|\unip\unir\verb|]|\uniu\univ\verb|[|\unii\uniG\verb|]|\verb|]| & {\em proving}\\
  718 & \verb|[|\unib\uniIX\verb|[|\unik\uniAH\unim\verb|]|\verb|]| & {\em become}&
 1076 & \verb|[|\unip\verb|[|\unil\uniE\verb|]|\uniZ\unir\verb|]| & {\em pleasure}\\
\end{tabular}

\vspace{.25in}

\begin{tabular}{ll}\hline
\multicolumn{2}{l}{Cereal grains have been used for centuries to prepare fermented beverages.} \\
\hspace{.15in} & \unis\uniI\unir\unii\unil\unig\unir\unie\unii\unin\uniz\unih\uniE\univ\unib\unii\unin\uniy\uniu\uniz\unid\unif\unir\unis\uniE\unin\unit\unir\unii\uniz\unit\uniAX\unip\unir\unip\uniE\unir\unif\unir\unim\uniE\unin\uniIX\unid\unib\uniE\univ\unir\uniIX\unid\uniJ\uniIX\uniz {\em (trans.)}\\
 &
 \unis\verb|[|\uniI\unir\unii\verb|]|\unil\verb|[|\unig\unir\unie\unii\unit\unid\verb|]|\verb|[|\unih\uniE\univ\unir\unii\verb|]|\unin\verb|[|\uniy\uniu\uniz\unid\verb|]|\verb|[|\unif\unir\verb|]|\verb|[|\unis\uniE\unin\unit\unir\verb|]|\unii\uniz\verb|[|\unit\uniAX\verb|]|\verb|[|\unip\unir\verb|]|\verb|[|\unip\uniE\unir\verb|]|\verb|[|\unif\unir\verb|]|\verb|[|\unim\uniE\unin\unii\verb|]|\unid\verb|[|\unib\uniE\univ\unir\unii\unid\uniJ\verb|]|\verb|[|\uniIX\uniz\verb|]| \\ \cline{1-2}
\multicolumn{2}{l}{This group is secularist and their program tends to be technological.} \\
 & \uniD\uniI\unis\unig\unir\uniu\unip\uniI\unis\uniE\unik\uniy\uniIX\unil\unir\uniI\unis\unit\uniIX\unin\uniD\uniE\unir\unip\unir\unio\uniu\unig\unir\uniE\unim\unit\uniE\unin\uniz\unit\uniIX\unib\unii\unit\uniE\unik\unin\uniAX\unil\unia\unid\uniJ\uniIX\unik\unil {\em (trans.)}\\
 &
 \verb|[|\uniD\uniI\unis\verb|]|\verb|[|\unig\unir\uniu\unip\verb|]|\uniI\unis\verb|[|\uniE\unig\uniy\uniIX\verb|]|\unil\verb|[|\unir\uniI\unis\unit\verb|]|\verb|[|\uniIX\unin\verb|]|\verb|[|\uniD\uniE\unir\verb|]|\verb|[|\unip\unir\unio\uniu\unig\unir\uniA\unim\verb|]|\verb|[|\unit\uniE\unin\verb|]|\uniz\verb|[|\unit\uniIX\unib\unii\verb|]|\verb|[|\unit\uniE\unik\unis\verb|]|\verb|[|\uniAX\unil\unia\unid\uniJ\uniIX\unik\unil\verb|]| \\ \cline{1-2}
\multicolumn{2}{l}{A portable electric heater is advisable for shelters in cold climates.} \\
 &
 \uniAX\unip\uniO\unir\unit\uniAX\unib\unil\uniAX\unil\uniE\unik\unit\unir\uniIX\unik\unih\unii\unit\unir\uniIX\uniz\uniIX\unid\univ\unia\unii\uniz\uniIX\unib\unil\unif\unir\uniS\uniE\unil\unit\unir\uniz\unin\unik\unio\uniu\unil\uniEPI\unik\unil\unia\unii\unim\uniIX\unit\unis {\em (trans.)}\\
 & \uniAX\verb|[|\unip\uniO\unir\unit\verb|]|\verb|[|\uniAX\unib\unil\verb|]|\verb|[|\uniAX\unil\uniE\unik\unit\unir\uniIX\unik\verb|]|\verb|[|\unih\unii\verb|]|\unit\unir\verb|[|\uniIX\uniz\verb|]|\verb|[|\uniIX\unid\univ\unia\unii\uniz\verb|]|\verb|[|\uniIX\unib\unil\verb|]|\verb|[|\unif\unir\verb|]|\verb|[|\uniS\uniE\unil\unit\unir\verb|]|\uniz\unin\unik\verb|[|\unio\uniu\unil\verb|]|\uniEPI\unik\verb|[|\unil\unia\unii\verb|]|\verb|[|\unim\uniIX\unit\verb|]|\unis 

\end{tabular}
\end{center}

\end{small}
\caption{\label{fig:sdicts}Excerpts from the dictionaries after processing
  TIMIT data, and three segmentations of the transcripts.}
\pageline
\end{figure}

The final dictionary contains 1097 words after training on the
transcriptions, and 728 words after training on the speech.  Most of
the difference is in the longer words: as might be
expected, performance is much poorer on the raw speech.
Figure~\ref{fig:sdicts} contains excerpts from both dictionaries and
several segmentations of the transcriptions.  Except for isolated
sentences, the segmentations of the speech data are not particularly
impressive.

Despite the relatively small size of the dictionary learned from raw
speech, we are very happy with these results. Obviously, there are real
limits to what can be extracted from a short, noisy data set.  Yet our
algorithm has learned a number of long words well enough that they can
be reliably found in the data, even when the underlying form does not
exactly match the observed input.  In many cases (witness {\em sometime},
{\em maybe}, {\em set aside} in figure~\ref{fig:sdicts}) these words are
naturally and properly represented in terms of others.  We expect that
performance will improve significantly on a longer, more regular corpus.
Furthermore, as will be described in the next section, the algorithm can be
extended to make use of side information, which has been shown to make the
word learning problem enormously easier.

\section{Extensions}\label{extensions}

Words are more than just sounds-- they have meanings and syntactic roles,
that can be learned using very similar techniques to those we have
already described.  Here we very briefly sketch what such extensions
might look like.

\subsection{Word Meanings}

An implicit assumption throughout this paper is that sound is learned
independently of meaning.  In the case of child language acquisition this
is plainly absurd, and even in engineering applications the motivation for
learning a sound pattern is usually to pair it with something else (text,
if one is building a dictation device, or words from another language in
the case of machine translation).  The constraint that the meaning of a
sentence places on the words in it makes learning sound and meaning
together much easier than learning sound alone
(Siskind~\cite{Siskind92,Siskind93b,Siskind94}, de
Marcken~\cite{deMarcken94b}).

Let us make the naive assumption that meanings are merely sets (the meaning
of /\uniD\uniAX/ might be $\{ t, h, e \}$, or the meaning of {\em temps
  perdu } might be $\{ \mbox{\em past}, \mbox{\em times}
\}$)\footnote{Siskind \cite{Siskind94} argues that it is easy to extend a
  program that learns sets to learn structures over them.}. If the meaning
of a sentence is a function of the meanings of the words in the sentence
(such as the union), then the meaning of a word
should likewise be that function applied to the meanings of the words in its
representation.  There must be some way to modify this default behavior,
such as by writing down elements that occur in a word's meaning but not
in the meanings of its representation and {\it vice versa}.

Assume that with each utterance $u$ comes a distribution over meaning sets,
$f(\cdot)$, that reflects the learner's prior assumptions about the meaning
of the utterance.  This side information can be used to improve
compression.  The learner first selects whether or not to make use of the
distribution over meanings (in this way it retains the ability to encode
the unexpected, such as {\em colorless green ideas sleep furiously}).  If
meaning is used, then $u$ is encoded under $p_G(u|f)$ rather than $p_G(u)$.
As an example of why this helps, imagine a situation where $f(M) = 0 \mbox{
  if } m\in M$.  Then since there is no chance of a word with meaning $m$
occurring in the input, all words with that meaning can effectively be
removed from the dictionary and probabilities renormalized.  The
probabilities of all other words will increase, and their code lengths
shorten.  Since word meanings are tied to compression, they can be
learned by altering the meaning of a word when such a move reduces the
combined description length.

We have fleshed out this extension more fully and conducted some initial
experiments, and the algorithm seems to learn word meanings quite reliably
even given substantial noise and ambiguity.  At this time, we have not
conducted experiments on learning word meanings with speech, though the
possibility of learning a complete dictation device from speech and
textual transcripts is not beyond imagination.

\subsection{Surface Syntax}

One of the major inadequacies of the simple concatenative language model we
have used here is that it makes no use of context, and as a consequence
grammars are much bigger than they need to be, and do not generalize as
well as they could.  The algorithm also occasionally produces words that
cross real-word boundaries, like \verb*|ed by the| (see
figure~\ref{fig:words}).  This is an example of a regularity that arises
because of several interacting linguistic processes that the algorithm can
not capture because it has no notion of abstract categories.  We would
prefer to capture this pattern using a form more like \verb|<|{\em
  vroot}\verb*|>ed by the<|{\em noun}\verb|>|, with an internal
representation that conformed to our linguistic intuitions.  But before we
can admit surface forms with underspecified categories, there must be some
means of ``filling in'' these categories.  Happily, this can be done
without ever leaving the concatenative framework, by representing tree
structure with extended left derivation strings.  Notice that \verb|<|{\em
  vroot}\verb*|>ed by the<|{\em noun}\verb|>| is the fringe of the parse
tree

\verb|[|$_{\mbox{\em vp}}$\verb|[|$_{\mbox{\em verb}}$\verb|<|{\em vroot}\verb*|>ed][|$_{\mbox{\em pp}}$\verb*|[ by][|$_{\mbox{\em np}}$\verb*|[ the]<|{\em noun}\verb|>]]]|.

\noindent This tree can be represented by the left derivation string

\verb|[|$_{\mbox{\em vp}}$\verb|<|{\em verb}\verb|><|{\em pp}\verb|>][|$_{\mbox{\em verb}}$\verb|<|{\em vroot}\verb|>ed]|$\diamond$\verb|[|$_{\mbox{\em pp}}$\verb*|[ by]<|{\em np}\verb|>][|$_{\mbox{\em np}}$\verb*|[ the]<|{\em noun}\verb|>]|$\diamond$

\noindent where the symbol $\diamond$ indicates that an underspecified
category like {\em vroot} is not expanded.  Thus, we have a means of
representing sequences of terminals and abstract categories by
concatenating context-free rules that look very much like words.  This
suggests merging the notion of rule and word.  Then words are sequences of
terminals and abstract categories that are represented by concatenating
words and $\diamond$'s.  The only significant differences between this
model and our current one is that words are linked to categories, and that
there must be some mechanism for creating abstract categories.  The first
difference disappears if each word has its own category; in essence, the
category takes the place of the word index.  In fact, if the notion of a
category replaces that of a word, then the representation of a category is
now a sequence of categories (and $\diamond$'s).  At this point, the only
remaining hurdle is the creation of abstract categories (which represent
sets of other categories).\footnote{This is not quite true; depending on
  the interpretation of the probability of a category given an abstract
  category, probability estimation procedures can change dramatically.} We
will not explore this further here.

Although we are just beginning work in this area, this close link between our
current representation and context-free grammars gives us great hope that
we can learn CFG's that compete with or better the best Markov models for
prediction problems, and produce plausible phrase structures.

\subsection{Applications}

The algorithm we have described for learning words has several properties
that make it a particularly good tool for solving language engineering
problems.  First of all, it reliably reproduces linguistic structure in its
internal representations.  This can not be said of most language models,
which are context based.  Using our algorithm for text compression, for
instance, enables the {\em compressed} text to be searched or indexed in
terms of intuitive units like words.  Together with the fact that the algorithm
compresses text extremely well (and has a rapid decompression counterpart),
this means it should be useful for off-line compression of databases
like encyclopedias.

Secondly, the algorithm is unsupervised.  It can be used to construct
dictionaries and extend existing ones without expensive human labeling
efforts.  This is valuable for machine translation and text indexing
applications.  Perhaps more importantly, because the algorithm constructs
dictionaries from observed data, its entries are optimized for the
application at hand; these sorts of dictionaries should be significantly
better for speech recognition applications than manually constructed ones
that do not necessarily reflect common usage, and do not adapt themselves
across word boundaries ({\it i.e.}~no {\em wanna} like words).

Finally, the multi-layer lexical representation used in the algorithm is
well suited for tasks like machine translation, where
idiomatic sequences must be represented independently of the words they are
built from, while at the same time the majority of common sequences
function quite similarly to the composition of their components.

\section{Related Work}\label{related}

This paper has touched on too many areas of language and induction to
present an adequate survey of related work here.  Nevertheless, it is
important to put this work in context.

The use of compression and prediction frameworks for language induction is
quite common; Chomsky~\cite{Chomsky55} discussed them long ago and notable
early advocates include Solomonoff~\cite{Solomonoff60}.
Olivier~\cite{Olivier68} and Wolff~\cite{Wolff80,Wolff82} were among the
first who implemented algorithms that attempt to learn words from text
using techniques based on prediction.  Olivier's work is particularly
impressive and very similar to practical dictionary-based compression
schemes like LZ78~\cite{Ziv78}.  More recent work on lexical acquisition
that explicitly acknowledges the cost of parameters includes
Ellison~\cite{Ellison92} and Brent~\cite{Brent93a,Brent93b,Cartwright94}.
Ellison has used three-level compression schemes to acquire intermediate
representations in a manner similar to how we acquire words.  Our
contributions to this line of research include the idea of recursively
searching for regularities within words and the explicit interpretation of
hierarchical structures as a linguistic representation with the possibility
of attaching information at each layer.  Our search algorithm is also more
efficient and has a wider range of transformations available to it than
other schemes we know of, though such work as Chen~\cite{Chen95} and
Stolcke~\cite{Stolcke94} use conceptually similar search strategies for
grammar induction.  Recursive dictionaries themselves are not new, however:
a variety universal compression schemes (such as LZ78) are based on this
idea.  These schemes use simple on-line strategies to build such
representations, and do not perform the optimization necessary to arrive at
linguistically meaningful dictionaries.  An attractive alternative to the
concatenative language models used by all the researchers mentioned here is
described by Della Pietra {\em et al}~\cite{DellaPietra95}.

The unsupervised acquisition of words from continuous speech has received
relatively little study.  In the child psychology literature there is
extensive analysis of what sounds are available to the infant (see Jusczyk
\cite{Jusczyk93,Jusczyk94}), but no emphasis on testable theories of the
actual acquisition process.  The speech recognition community has generally
assumed that segmentations of the input are available for the early stages
of training.  As far as we are aware, our work is the first to use a model
of noise and phonetic variation to link speech to the sorts of learning
algorithms mentioned in the previous paragraph, and the first attempt to
actually learn from raw speech.

\section{Conclusions}\label{conclusions}

We have presented a general framework for lexical induction based on a form
of recursive compression.  The power of that framework is demonstrated by
the first computer program to acquire a significant dictionary from raw
speech, under extremely difficult conditions, with no help or prior
language-specific knowledge.  This is the first work to present a complete
specification of an unsupervised algorithm that learns words from speech,
and we hope it will lead researchers to study unsupervised
language-learning techniques in greater detail.  The fundamental simplicity
of our technique makes it easy to extend, and we have hinted at how it can
be used to learn word meanings and syntax.  The generality of our algorithm
makes it a valuable tool for language engineering tasks ranging from the
construction of speech recognizers to machine translation.

The success of this work raises the possibility that child
language acquisition is {\em not} dependent on supervisory clues in the
environment.  It also shows that linguistic structure can be extracted from
data using statistical techniques, if sufficient attention is paid to the
nature of the language production process.  We hope that our results can be
improved further by incorporating more accurate models of morphology and
phonology.

\subsection*{Acknowledgments}

The author would like to thank Marina Meil\u{a}, Robert Berwick, David
Baggett, Morris Halle, Charles Isbell, Gina Levow, Oded Maron, David
Pesetsky and Robert Thomas for discussions and contributions related to
this work.

\begin{small}

\end{small}

\appendix
\section{Phonetic Model}

This appendix presents the full set of phonemes (and identically, phones)
that are used in the experiments described in section~\ref{speechhmms},
and in examples in the text.  Each phoneme is a bundle of specific values
for a set of features.  The features and their possible values are also
listed below; the particular division of features and values is somewhat
unorthodox, largely because of implementation issues.  If a feature is
unspecified for a phoneme, it is because that feature is (usually for
physiological reasons) meaningless or redundant given other settings: {\em
  reduced} and {\em high} are defined only for vowels; {\em low}, {\em
  back} and {\em round} are defined only for unreduced vowels; {\em ATR} is
defined only for unreduced vowels that are +high or -back, -low.  {\em
  continuant} is defined only for consonants; {\em articulator} for all
consonants except laterals and rhotics; {\em nasality} is defined only for
stops; {\em sonority} is defined only for sonorants; {\em anterior} and
{\em distributed} are defined only for coronals; {\em voicing} is defined
only for non-sonorant, non-nasal consonants and laryngeals.  See
Kenstowicz~\cite{Kenstowicz94} for an introduction to such feature models.

\begin{small}
\def\ipa{\ipatenrm}
\begin{center}
\begin{tabular}{cll|cll}
Symbol & Example & Features & Symbol & Example & Features \\ \hline
\unib&\underline{b}ee&C,stop,lab,-n,-v&\unih&\underline{h}ay&laryngeal,-v\\
\unip&\underline{p}ea&C,stop,lab,-n,+v&\uniH&a\underline{h}ead&laryngeal,+v\\
\unid&\underline{d}ay&C,stop,cor,-n,-v,+a,-d&\uniI&b\underline{i}t&V,full,+h,-l,-b,-r,-ATR\\
\unit&\underline{t}ea&C,stop,cor,-n,+v,+a,-d&\unii&b\underline{ee}t&V,full,+h,-l,-b,-r,+ATR\\
\unig&\underline{g}ay&C,stop,dors,-n,-v&\uniU&b\underline{oo}k&V,full,+h,-l,+b,+r,-ATR\\
\unik&\underline{k}ey&C,stop,dors,-n,+v&\uniu&b\underline{oo}t&V,full,+h,-l+b,+r,+ATR\\
\uniJ&\underline{j}oke&C,fric,cor,-v,-a,-d&\uniE&b\underline{e}t&V,full,-h,-l,-b,-r,-ATR\\
\uniC&\underline{ch}oke&C,fric,cor,+v,-a,-d&\unie&b\underline{a}se&V,full,-h-l,-b,-r,+ATR\\
\unis&\underline{s}ea&C,fric,cor,-v,+a,-d&\uniAH&b\underline{u}t&V,full,-h,-l,+b,-r\\
\uniS&\underline{sh}e&C,fric,cor,-v,-a,+d&\unio&b\underline{o}ne&V,full,-h,-l,+b,+r\\
\uniz&\underline{z}one&C,fric,cor,+v,+a,-d&\uniA&b\underline{a}t&V,full,-h,+l,-b,-r\\
\uniZ&a\underline{z}ure&C,fric,cor,+v,-a,+d&\unia&b\underline{o}b&V,full,-h,+l,+b,-r\\
\unif&\underline{f}in&C,fric,lab,-v&\uniO&b\underline{ou}ght&V,full,-h,+l,+b,+r\\
\univ&\underline{v}an &C,fric,lab,+v&\uniIX&ros\underline{e}s&V,reduced,+h\\
\uniT&\underline{th}in&C,fric,cor,-v,+a,+d&\uniAX&\underline{a}bout&V,reduced,-h\\
\uniD&\underline{th}en&C,fric,cor,+v,+a,+d&\uniEPI&{\em silence} &silence\\
\unim&\underline{m}om &C,stop,lab,+n\\
\unin&\underline{n}oon&C,stop,cor,+n,+a,-d\\
\uniG&si\underline{ng}&C,stop,dors,+n\\
\unil&\underline{l}ay&C,sonorant,lateral\\
\unir&\underline{r}ay&C,sonorant,rhotic\\
\uniw&\underline{w}ay&C,sonorant,lab\\
\uniy&\underline{y}acht&C,sonorant,cor,+a,-d
\end{tabular}
\end{center}
\end{small}

\begin{small}
\begin{center}
\begin{tabular}{llll}
Feature & Values & $\mu$ & $\alpha$ \\ \hline
{\em consonantal} & silence, C (consonant), V (vowel), laryngeal & 0 & 0 \\
{\em continuant} & stop, fric (fricative), sonorant & 0.01 & 1 \\
{\em sonority} & lateral, rhotic, glide & 0 & 0 \\
{\em articulator} & lab (labial), cor (coronal), dors (dorsal) & 0 & 1 \\
{\em anterior} & +a (anterior), -a (not anterior) & 0.02 & 1 \\
{\em distributed} & +d (distributed), -d (not distributed) & 0.02 & 1 \\
{\em nasality} & +n (nasal), -n (non-nasal) & 0.01 & 1 \\
{\em voicing} & +v (voiced), -v (unvoiced) & 0.01 & 1 \\
{\em reduced} & reduced, full & 0.15 & 0 \\
{\em high} & +h (high), -h (not high) & 0.01 & 0 \\
{\em back} & +b (back), -b (front) & 0.01 & 0 \\
{\em low} & +l (low), -l (not low) & 0.01 & 0 \\
{\em round} & +r (rounded), -r (unrounded) & 0.01 & 0 \\
{\em ATR} & +ATR, -ATR & 0.01 & 0
\end{tabular}
\end{center}
\end{small}

\subsection*{Modeling the Generation of a Phone}

Both phonemes and phones are bundles of articulatory features (phonemes
representing the intended positions of articulators, phones the actual
positions).  As mentioned above, in certain cases the value of a feature
may be fixed or meaningless given the values of others.  We assume features
are generated independently: with $i$ ranging over free features, the
functions in figure~\ref{fig:transitions} can be written $p_I(s) = \prod_i
p^i_I(s^i)$, $p_M(s|q,u) = \prod_i p^i_M(s^i|q^i,u^i)$, and $p_C(s|q,u,n) =
\prod_i p^i_C(s^i|q^i,u^i,n^i)$.  Feature selection for insertion is under
a uniform distribution; for mapping and copying it depends on the
underlying phoneme and immediate context:

\vspace{-.2in}
\begin{eqnarray*}
p^i_I(s^i) & = & 1/Z^i.\\
p^i_M(s^i|q^i,u^i) & = & (\mu^i + \beta_u\delta(s^i,u^i)+ \alpha^i
\beta_q\delta(s^i,q^i))/Z^i(q^i,u^i).\\ 
p^i_C(s^i|q^i,u^i,n^i) & = & (\mu^i + \beta_u\delta(s^i,u^i) + \alpha^i
\beta_q \delta(s^i,q^i) + \alpha^i \beta_n
\delta(s^i,q^i))/Z^i(q^i,u^i,n^i).
\end{eqnarray*}
\vspace{-.2in}

\noindent Here, the $Z$'s are normalization terms; $\alpha^i$ is a 0-1
coefficient that determines whether a feature assimilates; $\mu^i$
determines the amount of noise in the mapping (generally in the range of
$0.01$ but as high as $0.15$ for the vowel-reduction feature); and
$\beta_q$ and $\beta_n$ determine the relative weighting of the input
features (both are equal to $0.15$ in the experiments we describe in this
paper).  Values for $\alpha^i$ and $\mu^i$ can be found in the chart above.

\section{Adding and Deleting Words}

This appendix is a brief overview of how the approximate change in
description length of adding or deleting a word is computed.  Consider the
addition of a word $X$ to the grammar $G$ (creating $G'$).  $X$ represents
a sequence of characters and will take the place of some other set of words
used in the representation of $X$.  Assume that the count of all other
words remains the same under $G'$.  Let $c(w)$ be the count of a word $w$
under $G$ and $c'(w)$ be the count under $G'$.  Denote the expected number
of times under $G$ that the word $w$ occurs in the representation of $X$ by
$c(w|X)$, and the same quantity under $G'$ with $c'(w|X)$.  Finally, let $C
= \sum c(w)$ and $C' = \sum c'(w)$.  Then

\vspace{-.15in}
\begin{eqnarray*}
c'(w) & \approx & c(w) + c'(w|X) - c'(X)c(w|X).\\
p'(w) & = & \frac{c'(w)}{C'} \approx \frac{c(w) + c'(w|X) - c'(X)c(w|X)}{C +
  \left( \sum c'(w|X) \right) + c'(X)\left( 1- \sum c(w|X) \right)}.\\
p'(X) & = & \frac{c'(X)}{C'} \approx \frac{c'(X)}{C +
  \left( \sum c'(w|X) \right) + c'(X)\left( 1- \sum c(w|X) \right)}.
\end{eqnarray*}
\vspace{-.15in}

\noindent To compute these values, estimates of $c'(X)$ and $c'(w|X)$ must
be available (these are not discussed further).  These equations give
approximate values for probabilities and counts after the change is made.
The total change in description length from $G$ to $G'$ is given by

\vspace{-.15in}
\[ \Delta \approx -c'(X)\log p'(X) + \sum_{w} \left(
  -c'(w)\log p'(w) - -c(w)\log p(w) \right). \]
\vspace{-.15in}

\noindent This equation accounts for changing numbers and lengths of word
indices.  It is only a rough approximation, accurate if the Viterbi
analysis of an utterance dominates the total probability.  If $\Delta < 0$
then $X$ is added to the grammar $G$.  An important second order effect
(not discussed here) that must also be considered in the computation of
$\Delta$ is the possible subsequent deletion of components of $X$.  Many
words are added simultaneously: this violates some of the assumptions made
in the above approximations, but unnecessary words can always be deleted.

If a word $X$ is deleted from $G$ (creating $G'$) then in all places $X$
occurs words from its representation must be used to replace it.  This
leads to the estimates

\vspace{-.15in}
\begin{eqnarray*}
c'(X) & = & 0.\\
c'(w) & \approx & c(w) + c(w|X)(c(X)-1).\\
p'(w) & = & \frac{c'(w)}{C'} \approx
\frac{c(w) + c(w|X)(c(X)-1)}{C - c(X) + (\sum c(w|X))(c(X)-1)}.\\
\Delta & \approx & - -c(X)\log p(X) + \sum_{w} \left(
  -c'(w)\log p'(w) - -c(w)\log p(w) \right).
\end{eqnarray*}
\vspace{-.15in}

\noindent Again, $X$ is deleted if $\Delta < 0$.  Any error can be fixed in
the next round of word creation, though it can improve performance to avoid
deleting any nonterminal whose representation length has increased
dramatically as a result of other deletions.

\end{document}